%% file: main.tex
\documentclass[10pt,twocolumn,letterpaper]{article}

\usepackage{cvpr}              
\usepackage{graphicx}
\usepackage{booktabs}
\usepackage{xcolor}
\PassOptionsToPackage{table}{xcolor}
\usepackage{colortbl}
\usepackage{subcaption}
\usepackage{multicol}
\usepackage{multirow}
\usepackage{tcolorbox}
\usepackage{mathtools}
\usepackage{bm}
\tcbuselibrary{breakable, skins}
\usepackage[T1]{fontenc}
\usepackage{amsmath, amssymb, amsthm}
\newtheorem{definition}{Definition}
\usepackage{geometry}
\usepackage{enumitem}
\newcommand{\ProtoOS}{Planner-Atomic}
\newcommand{\ProtoOSeq}{Planner-Reactive}
\newcommand{\ProtoSW}{Planner-Managed}
\newcommand{\Best}[1]{\cellcolor[RGB]{220,240,220}#1}

\definecolor{cvprblue}{rgb}{0.21,0.49,0.74}
\usepackage[pagebackref,breaklinks,colorlinks,allcolors=cvprblue]{hyperref}

\title{How Far Are Vision-Language Models from Constructing the Real World? A Benchmark for Physical Generative Reasoning} 

\author{
Luyu Yang~~~~~Yutong Dai~~~~~An Yan~~~~~Viraj Prabhu~~~~~Ran Xu~~~~~Zeyuan Chen~\\[4pt]
Salesforce AI Research
}

\begin{document}

\twocolumn[{%
\renewcommand\twocolumn[1][]{#1}%
\maketitle
\begin{center}
    \centering
    \captionsetup{type=figure}
    \includegraphics[width=1.0\linewidth]{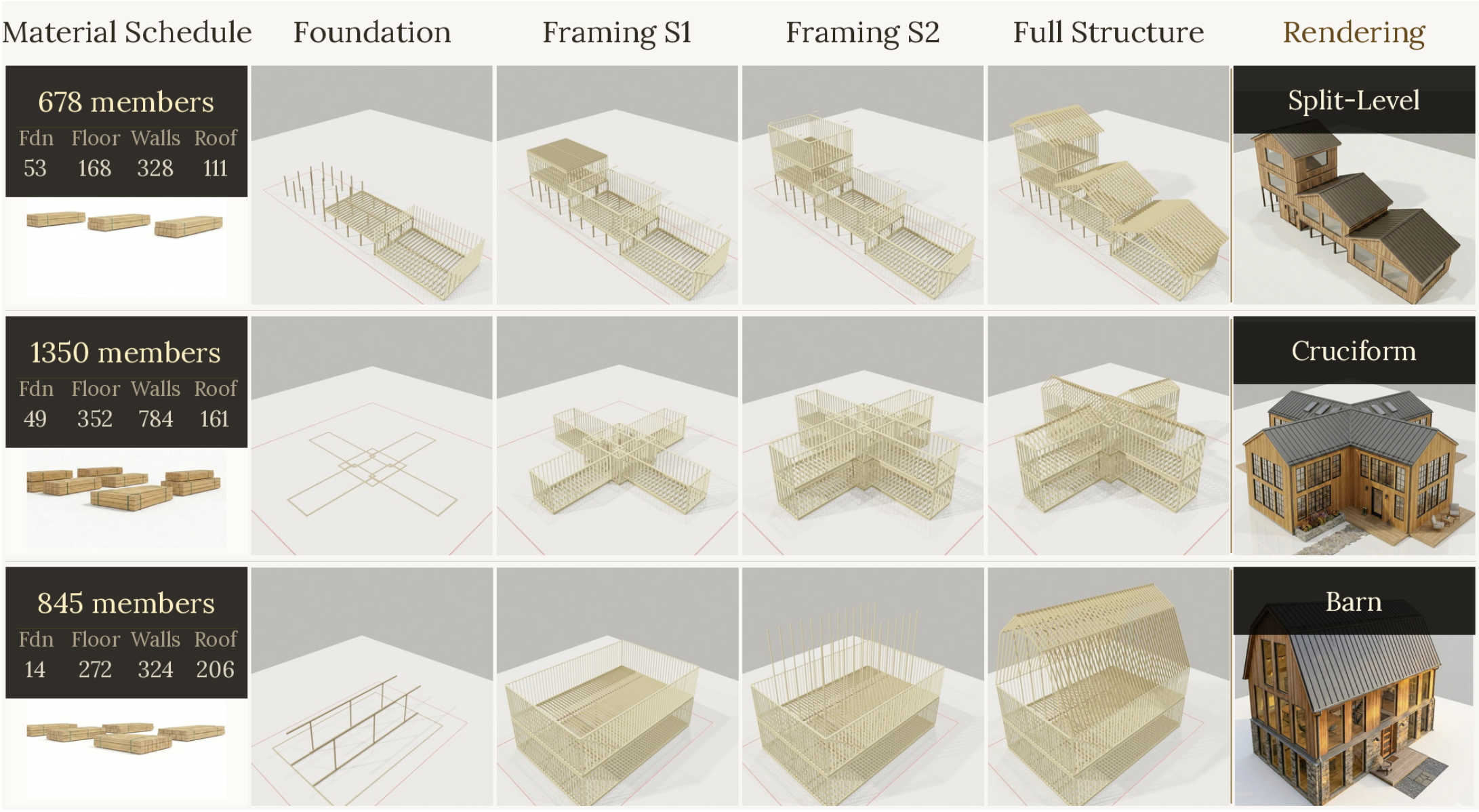}
    \captionof{figure}{\small\textbf{DreamHouse benchmark samples.}
Each row shows a single structure across five representations:
material schedule (member counts by subsystem), foundation, two intermediate
framing stages, and complete timber frame, alongside the target exterior rendering
from which the model must infer the hidden structural system.
Three of the 13 architectural styles are shown: Split-Level (678 members),
Cruciform (1{,}350 members), and Barn (845 members).
Member counts span four subsystems: foundation (Fdn), floor, walls, and roof.}
    \label{fig:fig_1_benchmark}
\end{center}
}]

\begin{abstract}
\input{sections/abstract}
\end{abstract}

\input{sections/intro}

\input{sections/related}

\input{sections/method}

\input{sections/experiments}

\input{sections/conclusion}

\input{appendix}

{
    \small
    \bibliographystyle{ieeenat_fullname}
    \bibliography{main}
}

\end{document}

%% file: sections/abstract.tex
The physical world is not merely visual; it is governed by rigorous structural and procedural constraints. Yet, the evaluation of vision-language models (VLMs) remains heavily skewed toward perceptual realism, prioritizing the generation of visually plausible 3D layouts, shapes, and appearances. Current benchmarks rarely test whether models grasp the step-by-step processes and physical dependencies required to actually \emph{build} these artifacts---a capability essential for automating design-to-construction pipelines. To address this, we introduce \textbf{DreamHouse}, a novel benchmark for \emph{physical generative reasoning}: the capacity to synthesize artifacts that concurrently satisfy geometric, structural, constructability, and code-compliance constraints. We ground this benchmark in residential timber-frame construction, a domain with fully codified engineering standards and objectively verifiable correctness. We curate over 26{,}000 structures spanning 13 architectural styles---each verified to construction-document standards (LOD\,350)---and develop a deterministic 10-test structural validation framework. Unlike static benchmarks that assess only final outputs, DreamHouse supports iterative agentic interaction. Models observe intermediate build states, generate construction actions, and receive structured environmental feedback, enabling a fine-grained evaluation of planning, structural reasoning, and self-correction. Extensive experiments with state-of-the-art VLMs reveal substantial capability gaps that are largely invisible on existing leaderboards. These findings establish physical validity as a critical evaluation axis orthogonal to visual realism, highlighting physical generative reasoning as a distinct and underdeveloped frontier in multimodal intelligence. Available at \href{https://luluyuyuyang.github.io/dreamhouse}{\textit{https://luluyuyuyang.github.io/dreamhouse}}

%% file: sections/intro.tex

\section{Introduction}
\label{sec:intro}

The physical world is not merely a surface; it is a rigorous system of constraints~\cite{marin2025sciblend,zha2025enable,ouyang2025spacer,wu2025reinforcing,stogiannidis2025mind,chen2025spatial,pan2025metaspatial,tang2025lego}. 
A floor must bear load, a rafter must respect its allowable span, and a wall must continuously transfer force from roof to foundation. 
These are not aesthetic choices, but fundamental physical realities. 
Yet, the dominant evaluation paradigm for generative vision models treats the world as purely visual, asking only if a model can produce outputs that \emph{look}   `correct~\cite{wang2025embodiedgen,kong20253d,team2025hunyuanworld,duan2025worldscore,alhaija2025cosmos,ding2025understanding}. 
Whether those outputs could physically \emph{stand} is a question the field has largely left unasked.

While the recent surge in vision-language models (VLMs) has yielded remarkable perceptual capabilities, evaluating their grasp of physical laws remains in its infancy~\cite{wang2025embodiedgen,zhang2025open3d,rodionov2025floorplanqa,chow2501physbench,chen2025think}. 
Recent benchmarks like PhysBench~\cite{chow2501physbench} and VSI-Bench~\cite{chen2025think} probe physical reasoning and spatial recall, but they are fundamentally \emph{comprehension} tasks---the model observes, answers, and is scored. 
They do not evaluate whether a model can \emph{build}. 
Generating a physically realizable artifact from scratch, under strict engineering constraints and without visible ground truth, requires moving beyond passive observation. 
Between perceiving a structure and constructing one lies a critical gap that standard benchmarks fail to measure.

To bridge this gap, we introduce \textbf{DreamHouse}, a novel benchmark designed to evaluate \textbf{physical generative reasoning}---the capacity to synthesize artifacts that concurrently satisfy geometric, structural, and code-compliance constraints. 
We ground our benchmark in residential timber-frame construction. 
This domain provides fully codified correctness criteria and discrete, verifiable components, while remaining visually complex enough to prove that perceptual plausibility alone is insufficient. 
DreamHouse comprises over 26{,}000 structurally verified models spanning 13 architectural styles. 
Furthermore, we provide a suite of 10 deterministic, physics-based tests covering load paths, span limits, member connectivity, \textit{etc.}. 
Crucially, these tests operate directly on the scene graph, bypassing the need for computationally expensive simulations.

The DreamHouse task is formulated as an iterative generation process: given rendered views of a target structure, a model must generate Blender Python construction code, process structured validation feedback, and refine its output until all structural tests pass. 
We evaluate performance across three protocols with varying degrees of external scaffolding (\ProtoOS, \ProtoOSeq~and \ProtoSW) and two input conditions (bare framing visible, \textit{Frame}, versus occluded by finished cladding, \textit{Facade}). 
This framework serves as a controlled ablation study, isolating whether generation failures stem from weak spatial reasoning, deficient planning, or an inability to self-correct.

Our extensive evaluation of three frontier VLMs inclyding GPT-5~\cite{singh2025openai}, Gemini 3~\cite{gemini3flash} and Claude 4.5~\cite{claude2025opus} across more than 20{,}000 independent agentic tasks reveals striking limitations in current state-of-the-art models. 
Notably, models that excel on standard coding and reasoning leaderboards often struggle with physical generation; highly-ranked generalists frequently underperform compared to their peers when evaluated under our structured protocols. 
Even the most capable model achieves a joint pass rate of \textbf{merely 7.1\%}, successfully satisfying both structural validity and visual fidelity simultaneously. 
These findings demonstrate that physical generative reasoning is not a natural byproduct of general intelligence, but rather a distinct, underdeveloped capability axis demanding dedicated evaluation. Our contributions:
\begin{itemize}
  \item \textbf{Physical generative reasoning} as a novel VLM
        evaluation axis, distinct from perception, comprehension,
        and simulation-based benchmarks.
  \item \textbf{The DreamHouse benchmark}, comprising 26{,}000+ verified
        timber-frame structures across 13 architectural styles
        with multi-view renderings and construction phase-wise annotations.
  \item \textbf{A 10-test structural validation suite} that is deterministic
        and simulation-free, covering International Residential Code (IRC) compliance~\cite{code2018international}, physics,
        geometry, and fabrication details.
  \item \textbf{A three-protocol agentic evaluation framework},
        showing that scaffold design is as important as
        model selection for physical generation tasks.
\end{itemize}

%% file: sections/related.tex
\section{Related Work}
\label{sec:related}

\noindent\textbf{Physical and Spatial Benchmarking for VLMs.}
A growing body of work exposes systematic gaps between VLM semantic
fluency and physical-world grounding~\cite{jia2025omnispatial,zhang2025open3d,rodionov2025floorplanqa,pun2025generating,chow2501physbench,chen2025think}.
PhysBench~\cite{chow2501physbench} evaluates 75 VLMs on 10K
video-image-text entries spanning object properties, relationships, and
dynamics, finding that models excel at static recognition but fail on
Newtonian dynamics, a deficiency attributed to missing physical priors
rather than perceptual limits.
VSI-Bench~\cite{yang2025thinking} measures metric visual-spatial
intelligence (distances, sizes, directions) from egocentric video,
showing that standard chain-of-thought prompting \emph{degrades} spatial
estimation, models must instead build explicit cognitive maps.
Think-with-3D~\cite{chen2025think} endows VLMs with 3D geometric
imagination via latent alignment and RL-based spatial rewards, enabling
occluded geometry completion from sparse views.
These benchmarks evaluate \emph{passive} physical understanding:
answering questions about or reconstructing observed scenes.
DreamHouse targets the harder \emph{generative} gap, producing
structures that are themselves physically valid, without any reference
structure to recall.

\noindent\textbf{3D Scene Generation.}
Large-scale 3D generative models~\cite{wang2025embodiedgen,kong20253d,team2025hunyuanworld,duan2025worldscore,alhaija2025cosmos,li2025worldmodelbench,zhang2025matrix} optimize photometric or geometric
metrics against reference renderings, leaving structural validity
unmeasured.
TRELLIS~\cite{xiang2025structured} achieves state-of-the-art
image-conditioned 3D generation via sparse voxel latents, but its
representation is derived entirely from visual observations, a
generated house mesh may score well on Chamfer distance while failing
every structural test.
SpatialGen~\cite{fang2025spatialgen} generates photorealistic indoor
scenes conditioned on 3D layouts, operating at rendering-grade
level-of-detail for design visualization.
DreamHouse operates at fabrication-grade detail (LOD~350),
specifying member species, cross-sections, and connections sufficient
for construction scheduling, complementary positions in the AEC
pipeline.

\noindent\textbf{Code-Driven Structured Generation.}
A productive line of work~\cite{tang2025charts,zhou2025genco,doris2025cad,chen2025code2video,yin2025codediffuser,sun2025januscoder} has VLMs write executable programs verified
by domain-specific oracles, consistently outperforming direct
pixel-level synthesis for structured outputs.
DreamHouse adopts this architecture but replaces the visual oracle
with deterministic engineering compliance.
VIGA~\cite{yin2026vision} frames inverse graphics as a long-horizon
agentic loop (\textsc{write}$\to$\textsc{run}$\to$\textsc{render}$\to$%
\textsc{compare}$\to$\textsc{revise}), demonstrating that iterative
execution feedback recovers performance where single-shot VLMs fail.
Its feedback signal is photometric; structural failures (missing
members, span violations) are invisible to any rendering oracle.
BlenderGym~\cite{gu2025blendergym} benchmarks VLMs on Blender editing
tasks via LPIPS/CLIP-I, sharing our infrastructure but evaluating visual
imitation rather than structural correctness.
MCP-Universe~\cite{luo2025mcp} benchmarks LLM agents across
six domains including 3D Design, where even GPT-5 achieves only 43.7\%
success; DreamHouse provides the domain-specific 16-test physical
validation absent from its 3D tasks.
AutoPresent~\cite{ge2025autopresent} and
Chart2Code~\cite{tang2025charts} further confirm that programmatic
generation dominates pixel-level synthesis for structured outputs;
DreamHouse extends this finding to 3D structural engineering where
constraints are physical law rather than aesthetic convention.

\noindent\textbf{Physically-Grounded Assembly.}
The works closest to DreamHouse require agents to produce physical
assemblies whose validity is determined by physics, not appearance~\cite{lee2021ikea,wang2022ikea,ben2024ikea}.
BrickGPT~\cite{pun2025generating} is the most
directly analogous: it formulates LEGO assembly as next-token prediction
with physics-aware rollback, explicitly framing its contribution as
``buildable, not just renderable.''
The technical gap is substantial: LEGO involves isotropic material with
uniform stud connectivity and a single stability check (center-of-mass
within support polygon), whereas timber framing involves orthotropic
material, heterogeneous connection mechanics, and multi-condition
verification (load-path connectivity, section modulus compliance, IRC
span limits, assembly dependency ordering).
Beyond domain complexity, BrickGPT~\cite{pun2025generating} is a trained generation method;
DreamHouse is an evaluation benchmark.
Video2Policy~\cite{ye2025video2policy} instantiates the same
visual-to-executable-program paradigm for robotic manipulation,
representing a complementary instance of visual-to-physical program
synthesis in a distinct physical domain.

%% file: sections/method.tex

\section{DreamHouse Benchmark}
\label{sec:method}

\begin{figure*}[t]
\centering
\includegraphics[width=1.0\linewidth]{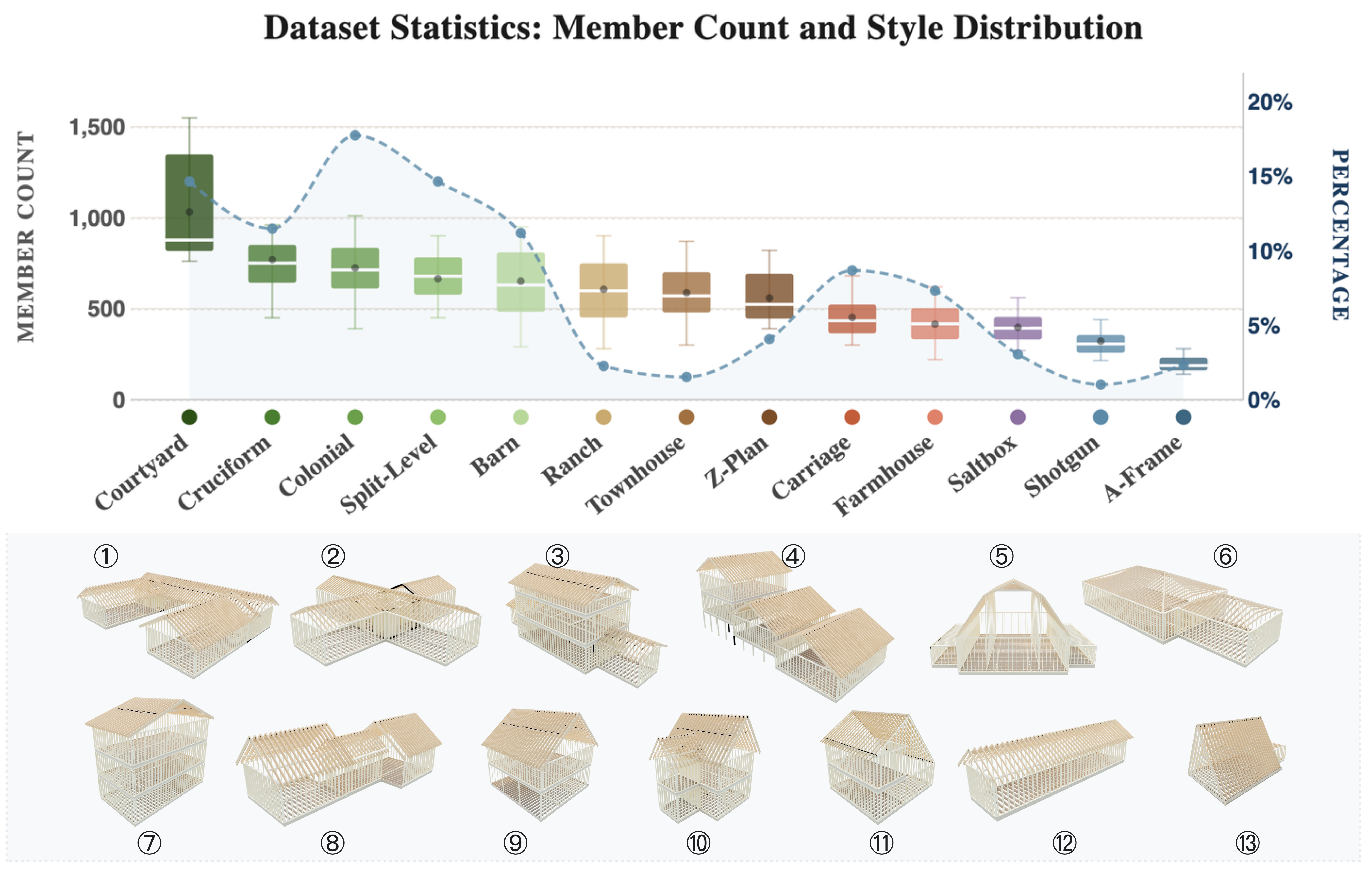}
\caption{\small\textbf{DreamHouse Dataset Overview.} \textit{(Top)} Dataset statistics showing member count distributions (box plots, left axis) and style proportions (dashed line, right axis) across all \textbf{13} architectural styles, ordered by decreasing structural complexity. Member counts range from 133 to 1,548 (mean 673). \textit{(Bottom)} Representative Cycles-rendered timber frame structures for each style: \textit{Courtyard, Cruciform, Colonial, Split-Level, Barn, Ranch, Townhouse, Z-Plan, Carriage, Farmhouse, Saltbox, Shotgun, A-Frame}, spanning complex multi-wing configurations to compact single-story forms.}
\label{fig:fig_6_dataset_distr}
\end{figure*}

\begin{figure*}[t]
    \centering
    \includegraphics[width=1.0\linewidth]{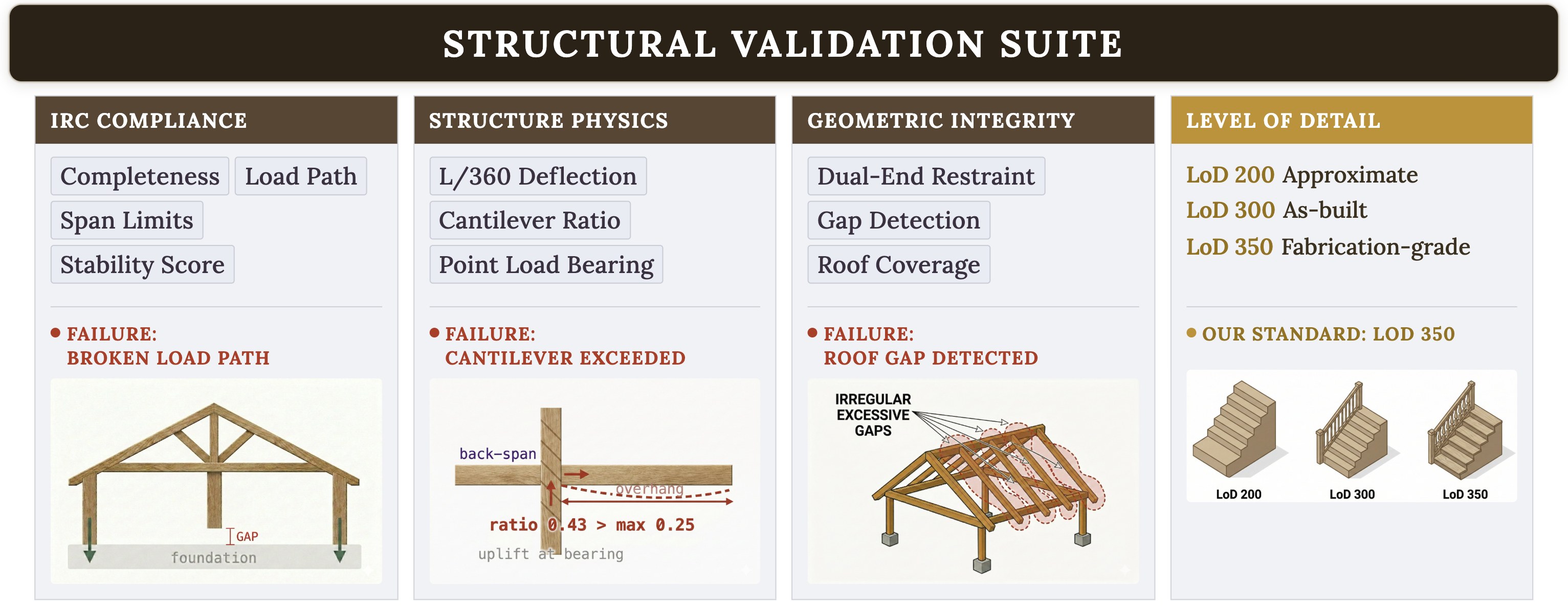}
    \caption{\small\textbf{Structural Validation Suite.}
    10 tests across four pillars. \textbf{IRC Compliance} (left): load path,
    span limits, completeness, stability score.
    \textbf{Structure Physics} (center-left): L/360 deflection, cantilever
    ratio, point load bearing.
    \textbf{Geometric Integrity} (center-right): dual-end
    restraint, gap detection, roof coverage.
    \textbf{LoD\,350} (right): fabrication-grade member geometry.}
    \label{fig:fig_3_validation_suite}
\end{figure*}

\begin{figure*}[t!]
    \centering
    \includegraphics[width=1.0\linewidth]{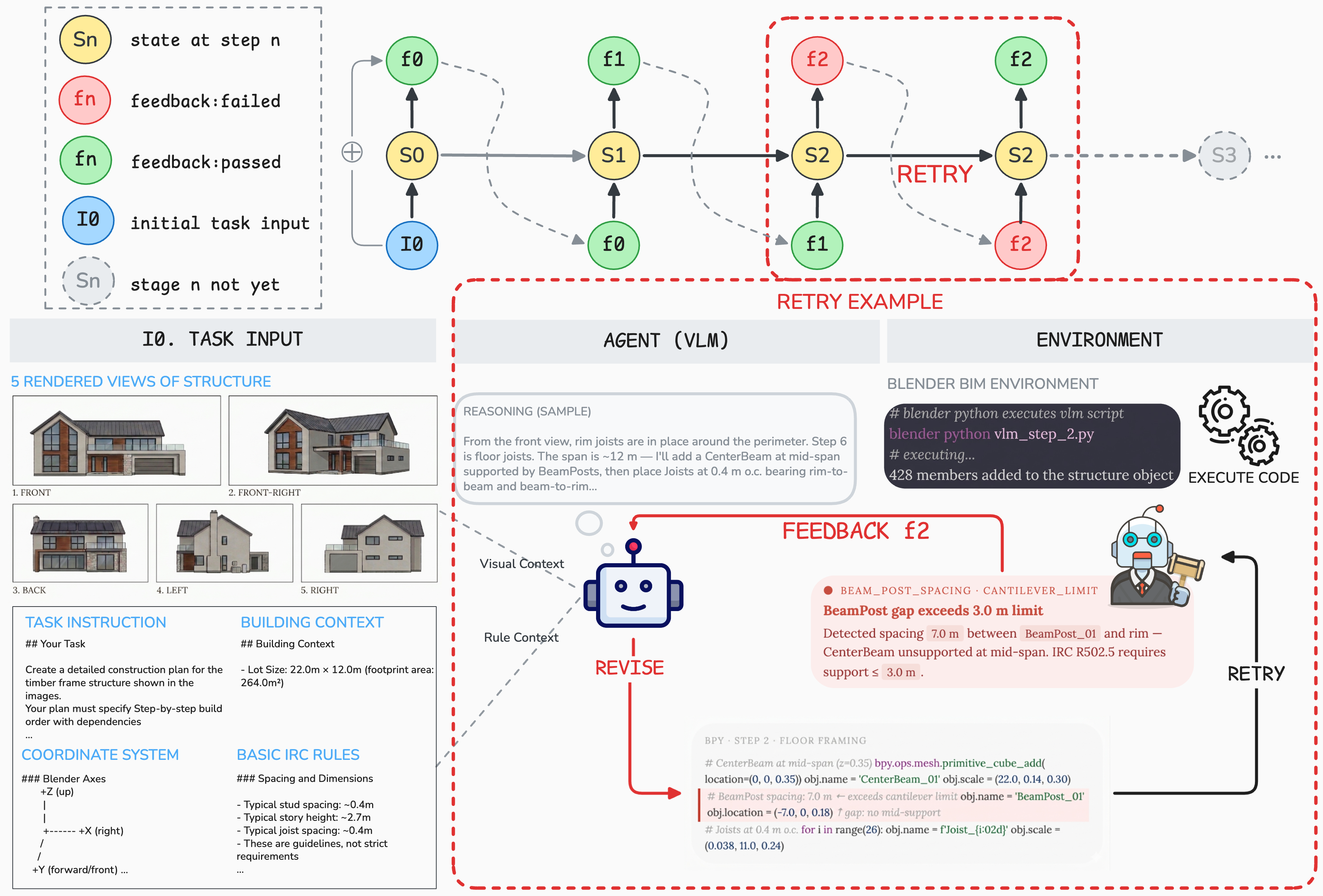}
    \caption{\small\textbf{Task formalization example \ProtoSW.}
\textit{Top (agent loop):} Evaluation is formalized as a recurrent agentic 
process. At each turn $t$, the agent $\mathcal{A}$ (VLM) receives observation 
$o_t = (I_0,\, f_{t-1})$, the original multi-view task image $I_0$ and 
structured validation feedback $f_{t-1}$ from the previous turn, and 
generates a Blender Python action $a_t$. The executor $\mathcal{E}$ applies 
$a_t$ to transition the scene graph $s_{t-1} \to s_t$, and the validator 
$\mathcal{V}$ produces feedback $f_t$ reporting per-test pass/fail and 
violation counts. This feedback becomes the input observation for the next 
turn, closing the loop. On failure (e.g., $f_2$, red), the agent retries from 
the same scene state $s_2$ without resetting context. 
$\mathcal{A}$, $\mathcal{E}$, and $\mathcal{V}$ are omitted from the diagram 
for visual clarity; see Section~\ref{sec:tasks} for full formalization.
\textit{Bottom (task instantiation):} The task input $I_0$ consists of five 
rendered views of the target structure paired with building context and 
rules. The agent reasons over these to produce and iteratively revise 
construction code; the environment executes the code in Blender and the 
validator returns structured diagnostic feedback driving the next revision.}
    \label{fig:fig_2_pipeline}
\end{figure*}

\begin{figure*}[t!]
\centering
\includegraphics[width=1.0\linewidth]{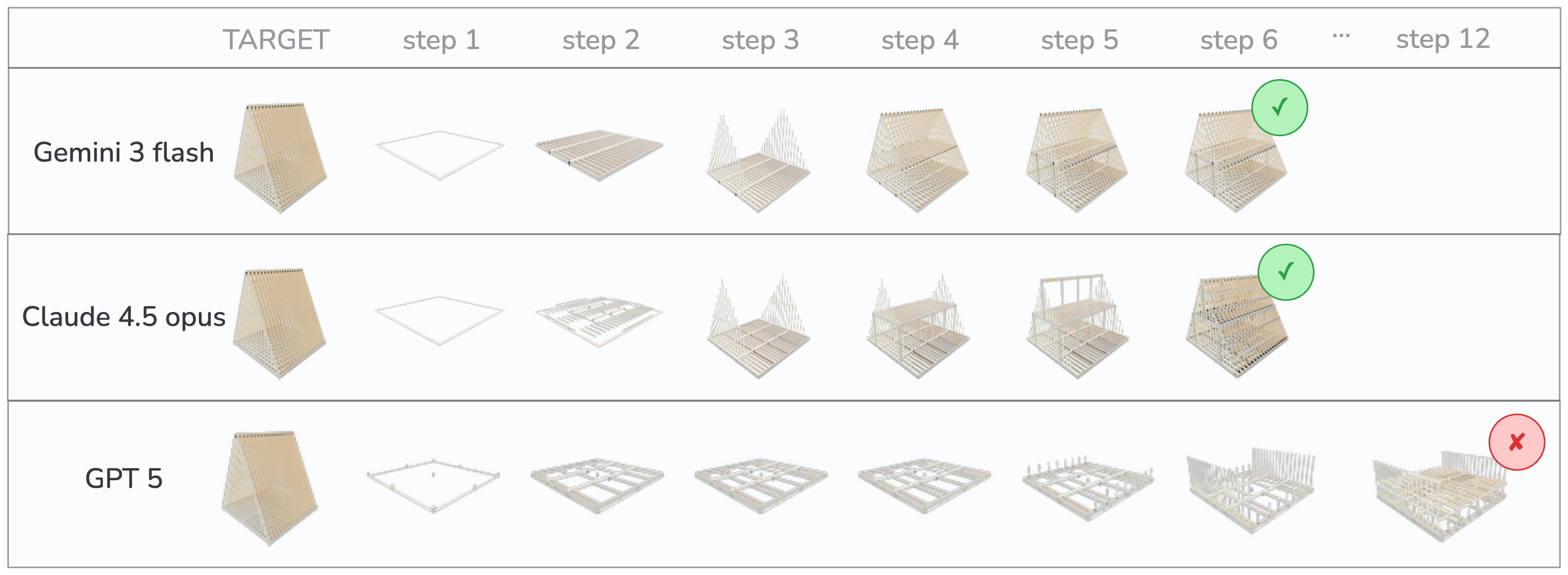}
\caption{\small\textbf{\ProtoSW~qualitative example AF-01-0060 (A-frame style).}
All three models begin from scratch and receive stepwise visual feedback toward the same target structure.
Despite reaching a valid result at step~6, \textbf{Gemini} and \textbf{Claude} employ markedly different construction strategies.
Gemini pursues a top-down, shape-first approach: it approximates the overall silhouette early and refines toward it.
Claude reasons bottom-up: it first establishes a structurally sound interior frame, then lays the roof rafters over it in the final step -- a sequence closer to how a builder would physically construct an A-frame.
\textbf{GPT-5} fails to recognize the defining constraint of A-frame geometry, that the roof planes double as load-bearing walls, and from step~6 onward enters a false loop of adding conventional wall studs along the perimeter.
Unable to escape this structural misconception, it exhausts all attempts without producing a valid result.
This example highlights that identical visual feedback can elicit fundamentally different reasoning strategies, and that success depends not just on visual matching ability but on implicit architectural knowledge.}
\label{fig:example_af_01_0060}
\end{figure*}

We construct \textbf{DreamHouse}, a benchmark for physical generative reasoning
grounded in residential timber-frame construction.
It comprises (1) a large-scale dataset of structurally verified structures
paired with multi-view renderings, and (2) a deterministic validation suite
of 10 physics-based tests scoring any generated structure against engineering
and code-compliance standards~\cite{code2018international,peng2024analysis}.
Together they define both the task and the metric, enabling objective evaluation
of whether a model can \emph{construct}, not merely depict, the physical world.

\subsection{Benchmark Construction}
\label{sec:dataset}

\paragraph{Testbed choice.}
Residential timber framing is an unusually well-suited domain for evaluating
physical generative reasoning: its correctness criteria are fully codified
(load-path integrity, member sizing, connection geometry, assembly order)
and objectively verifiable without human annotation~\cite{xia2014perceived}.
The domain is also visually rich and stylistically diverse, making perceptual
plausibility a meaningful but insufficient criterion.
Before this choice, we piloted an alternative \textbf{bridge-generation} testbed but found that frontier
VLMs could easily satisfy all validation tests by recovering a small set of geometric
parameters analytically from the image, bypassing structural understanding
entirely; \textit{see Appendix for details.}
Timber framing forecloses this shortcut: member spacing, connection hierarchy and
load-path topology are not directly legible from exterior
renders, and no analytic inversion exists.

\paragraph{Parametric generation and dataset scale.}
Each structure originates from a JSON configuration defining footprint
dimensions, story count, roof pitch, overhang depth.
A procedural Blender Python generator instantiates this into a fully resolved
3D timber-frame model with human in the loop --- individual foundation sills, floor joists, wall studs,
headers, ridge beams, rafters, and collar ties, each at the correct position,
orientation, and IRC cross-section, with member identities and parent-child
relationships preserved in a structured scene graph.

We identify \textbf{13 canonical residential styles}~\cite{kwon2016us} whose structural logic
is sufficiently distinct to stress different aspects of the reasoning pipeline:
\textit{A-Frame}, \textit{Barn}, \textit{Carriage}, \textit{Colonial},
\textit{Courtyard}, \textit{Cruciform}, \textit{Farmhouse}, \textit{Ranch},
\textit{Saltbox}, \textit{Shotgun}, \textit{Split-Level}, \textit{Townhouse},
and \textit{Z-Plan}.
These span single- and multi-story footprints, symmetric and asymmetric roofs,
and rectangular to complex multi-wing plans (Figure~\ref{fig:fig_6_dataset_distr} and Figure~\ref{fig:fig_1_benchmark}).
After filtering through the full validation suite, the released dataset contains
\textbf{26{,}543 provably buildable structures}, ranging from 133 to 1{,}548
members (mean 673, median 656), partitioned into Foundation, Floor, Walls,
and Roof categories~\cite{adel2018design}.
All structures are rendered from five canonical viewpoints using Blender~\cite{hess2013blender} Cycles.
All structures meet \textbf{LoD\,350} (fabrication-grade); \textit{see Appendix for details}.

\paragraph{Structural Validation Suite.}
The suite comprises 10 deterministic tests in three pillars operating directly
on the scene graph. With no physics simulation required, our evaluation
can be fast, deterministic, and interpretable (Figure~\ref{fig:fig_3_validation_suite});
\textit{full mathematical definitions are provided in Appendix}.

\subsection{Task Formalization}
\label{sec:tasks}

We formalize evaluation as a family of \emph{agentic generation tasks}.
Let $\mathcal{A}$ denote the agent (VLM), $\mathcal{E}$ the executor
(Blender environment), and $\mathcal{V}$ the structural validator.
At each turn $t$, $\mathcal{A}$ receives observation
$o_t = (I_0,\, f_{t-1})$, where $I_0$ is the fixed multi-view task
input and $f_{t-1}$ is structured validation feedback from the
previous turn, and produces a Blender Python action $a_t$.
$\mathcal{E}$ applies $a_t$ to transition the scene graph
$s_{t-1} \to s_t$, and $\mathcal{V}$ evaluates $s_t$ to produce
feedback $f_t$ reporting per-test pass/fail status and violation
counts (e.g., \textit{``BeamPost gap exceeds 3.0\,m; detected
spacing 7.0\,m between BeamPost\_01 and rim''}).
This feedback closes the loop: $f_t$ becomes part of $o_{t+1}$.
On failure, the agent retries from the same scene state $s_t$
without resetting context; the full conversation history
$\mathcal{H}_t = (I_0,\, a_1, f_1,\, \ldots,\, a_{t-1}, f_{t-1})$
is preserved across all retries.
Before writing any code, $\mathcal{A}$ produces a hierarchical JSON
plan specifying member categories, phase ordering, and assembly
dependencies.
Each task enforces a per-step retry budget $R_{\mathrm{step}}$ and
a global budget $R_{\mathrm{global}}$.

We instantiate three protocols that vary the degree of external
phase management, forming a controlled ablation over agentic
scaffold design.The same deterministic validation suite $\mathcal{V}$ used to certify
benchmark structures during dataset construction serves as the
evaluation signal during task execution, ensuring ground-truth
and assessment criteria are fully aligned.

\paragraph{\ProtoOS~($\mathcal{T}_S$).}
$\mathcal{A}$ receives $o_1 = (I_0, \varnothing)$ and generates a
complete construction script $a_1$ covering all member categories
in a single code block.
On failure, $a_t$ is regenerated from updated $\mathcal{H}_t$;
accepted when $\mathcal{V}(s_t)$ passes all 10 tests or
$R_{\mathrm{global}}$ is exhausted.
This protocol tests holistic single-pass structural synthesis with
no intermediate feedback between phases.

\paragraph{\ProtoOSeq~($\mathcal{T}_Q$).}
$\mathcal{A}$ generates a single script covering all $K$
construction phases in a self-determined order.
After each phase $k$ is materialized into $s_t$, it is evaluated
against $\mathcal{V}_{\mathrm{mid}} \subset \mathcal{V}$ (load path
and stability); failure triggers full script regeneration from
$\mathcal{H}_t$, invalidating all phases $k{+}1,\ldots,K$.
Unlike $\mathcal{T}_S$, the model must re-plan all phase
interdependencies in context after each failure --- combining the
demands of holistic planning and sequential commitment without
any external scaffolding.

\paragraph{\ProtoSW~($\mathcal{T}_W$).}
$\mathcal{A}$ generates scripts one phase at a time under external
phase management.
The scene $s_t$ persists across phases; each phase must pass
$\mathcal{V}_{\mathrm{mid}}$ before the next is unlocked, with up
to $R_{\mathrm{step}}$ retries per phase.
A failure at phase $k$ never invalidates $s_{t-1}$: only $a_t$ is
regenerated, leaving all prior scene state intact.
This protocol provides the strongest external scaffolding and
isolates each phase as an independent sub-task.

\paragraph{Iterative Refinement with Visual Feedback ($\mathcal{T}_E$).}
Beyond the core generation protocols, we introduce an exploratory editing task to test whether models can refine visual alignment without breaking an already valid structure. Starting from $s_T$ that passed $\mathcal{V}$ under $\mathcal{T}_S$
but achieved visual fidelity score $S < \tau$ (see Section~\ref{sec:metrics}), $\mathcal{A}$ iteratively
refines $s_t$ while preserving structural validity.
Feedback $f_t$ is augmented with a rendered side-by-side comparison
against the target; $\mathcal{V}$ is re-run at every iteration and
any $a_t$ causing structural regression is rejected even if $S$
improves.

\subsection{Evaluation Metrics}
\label{sec:metrics}
\paragraph{Structural Validity.}
This metric axis is binary: a structure either satisfies all tests
in $\mathcal{V}$ or it does not.
We deliberately avoid partial credit here.
Physical constraints are discontinuous by nature, a single
unrestrained member or one over-spanned joist renders the assembly
unsafe regardless of how well the rest is built, and a graded
score would obscure this harshness.
over all $N$ evaluated structures per condition.

\paragraph{Visual Fidelity.}
We measure pixel-level agreement between multi-view renders of
the generated and target structures (Eq.\eqref{eq:visual}).
Standard perceptual embeddings such as DINO or CLIP are ill-suited
here: they are trained to be invariant to the geometric and
positional differences that matter most in construction
(a wall shifted by 0.5\,m looks semantically similar but is
structurally wrong), and they provide no signal on missing or
misplaced members that happen to fall outside the salient region.
We instead use alpha-weighted MSE.
Let $\hat{R}_v$ and $R_v$ be the rendered and target images of timber frames at
view $v$, with union alpha mask
$m_v = \alpha^g_v \cup \alpha^r_v$ restricting comparison to
non-transparent regions of either structure.
The visual fidelity score is:
\newcommand{\Ev}[1]{\frac{\sum_p m_v(p)\,\|#1\|^2}{\sum_p m_v(p)}}

\begin{equation}
\label{eq:visual}
  S = \frac{1}{V}\sum_{v=1}^{V}
      \max\!\left(0,\; 1 - \lambda \cdot
      \frac{\sum_p m_v(p)\, e_v(p)}{\sum_p m_v(p)}
      \right)
\end{equation}
averaged over $V$ orthographic views.
The union mask penalizes missing geometry, a member absent in
the generated structure leaves unmatched non-transparent pixels
in the target without rewarding empty background agreement.
Visual fidelity is computed on structurally passed structures only.

\paragraph{Topological Fidelity.}
Visual similarity can be gamed: a model that generates the correct
silhouette but with wrong member counts, misplaced connections, or
incorrect spatial hierarchy will score well visually while
producing an unbuildable structure (Eq.\eqref{eq:topo}).
We therefore measure structural correspondence directly on the
scene graph, independent of rendering, via three complementary
statistics: Census accuracy $C$, Hungarian match rate $M$, and
Voxel IoU $V$.

\noindent$C$ measures whether the model generates the correct number of
members per category.
Let $n^*_k$ and $\hat{n}_k$ be the ground-truth and generated
member counts for category $k \in \{1,\ldots,K\}$; then
$C = \frac{1}{K}\sum_{k}\min(n^*_k, \hat{n}_k) / \max(n^*_k, \hat{n}_k)$.
This catches global over- or under-building even when individual
member positions look plausible, a failure mode that positional
metrics miss.

\noindent$M$ measures whether individual members are placed at the correct
spatial positions.
Let $\sigma^*$ be the optimal assignment from the Hungarian
algorithm between ground-truth and generated member centroids
$\{c_i\}$ and $\{\hat{c}_j\}$; then
$M = \frac{1}{N}\sum_{i} \mathbf{1}[\|c_i - \hat{c}_{\sigma^*(i)}\| \leq \delta]$
where $\delta = 0.3$\,m is the positional tolerance, chosen to
be permissive of minor placement error while rejecting members
shifted by more than a stud spacing.
$M$ catches misalignment even when counts are correct, a failure
mode $C$ cannot detect.

\noindent$V$ measures volumetric overlap of the assembled structure.
Let $\mathcal{G}$ and $\hat{\mathcal{G}}$ be the voxelized scene
graphs at a fixed grid resolution; then
$V = |\mathcal{G} \cap \hat{\mathcal{G}}| / |\mathcal{G} \cup \hat{\mathcal{G}}|$.
$V$ catches gross shape errors, \eg missing wings, wrong footprint
extent, collapsed roof geometry that member-level statistics
miss because they operate on individual centroids rather than
the assembled volume.

The composite score is:
\begin{equation}
\label{eq:topo}
  T = w_C\,C + w_M\,M + w_V\,V, \quad w_C + w_M + w_V = 1
\end{equation}
where $w_M > w_C = w_V$, reflecting that positional match is the
most direct measure of assembly correctness, with count accuracy
and volumetric overlap contributing equally as complementary
diagnostics.
Specific weight values are provided in Section~\ref{sec:experiments}.

%% file: sections/experiments.tex
\section{Experiments}
\label{sec:experiments}

\input{tables/results_table}
\input{tables/exp3_table}

\paragraph{Models.}
We evaluate three frontier vision-language models:
\textbf{GPT-5}~\cite{singh2025openai},
\textbf{Gemini~3~Flash}~\cite{gemini3flash},
\textbf{Claude~Opus~4.5}~\cite{claude2025opus}.
All models use identical prompting with no fine-tuning. We abbreviate Claude Opus~4.5 and Gemini~3~Flash as
\textbf{Claude} and \textbf{Gemini} respectively for brevity.

\paragraph{Evaluation subset.}
We evaluate on a stratified sample of 1{,}200 structures per
model-protocol cell, drawn uniformly across all 13 styles and 4 roof types, each structure represents a full agentic task: a multi-turn conversation of up to $\sim$50 API calls depending on protocol. Across three models and our three main protocols, this yields \textbf{21,600} independent agentic tasks and \textbf{$\sim$200k} API calls. Token consumption is correspondingly substantial, totalling approximately \textbf{$\sim$5B tokens} across the benchmark. This scale reflects the real computational cost of evaluating frontier models on physically grounded generative tasks.

\paragraph{Details.}
Visual fidelity uses $V=5$ orthographic views (front, front-right, back, left, right) at $512{\times}512$
with scale factor $\lambda=10$, mapping a 10\% mean pixel error
to $S=0$ and passing threshold $\tau=0.6$, which corresponds
empirically to structures visually recognizable as matching the
target style and layout.
Topological fidelity is computed with positional tolerance
$\delta=0.3$\,m (permissive of minor placement error while
rejecting members shifted by more than a stud spacing) and
composite weights $w_C=0.3$, $w_M=0.4$, $w_V=0.3$, reflecting
that positional match is the most direct measure of assembly
correctness (\textit{see Appendix for parameter choice}).
Both visual and topological scores are computed on structurally
passed structures only; the joint pass rate $J$ fraction
satisfying $\mathcal{V}$ and $S \geq \tau$ simultaneously
serves as the primary scalar summary of overall generation quality.

\subsection{Main Results}
\label{sec:results}
\begin{figure*}[t!]
\centering
\includegraphics[width=1.0\linewidth]{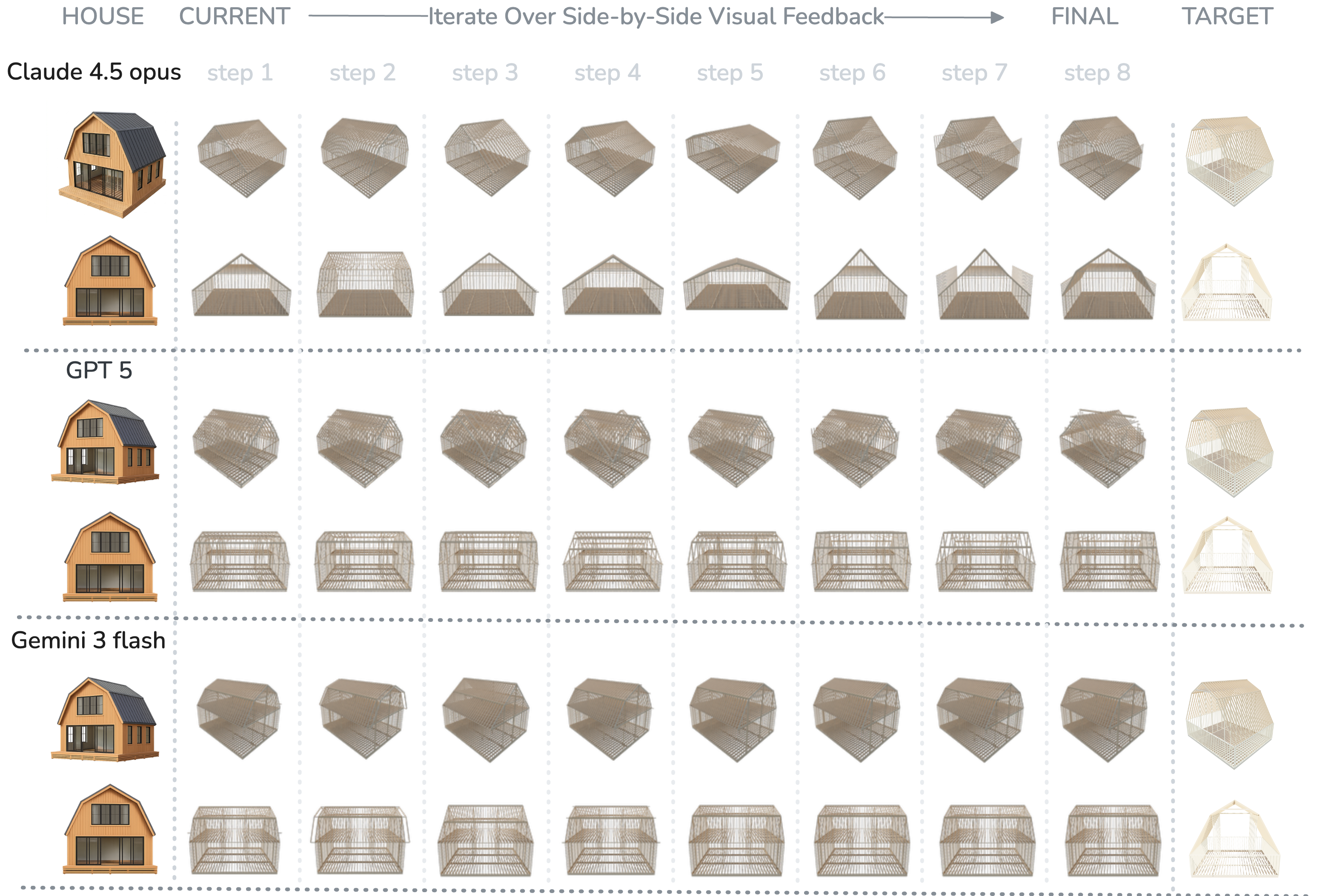}
\caption{\small\textbf{Iterative Refinement with Visual Feedback (Section ~\ref{sec:exp3}, qualitative example BN-01-0273 (barn style).}
Starting from a structurally valid \ProtoOS~output, each model receives up to 10 rounds of side-by-side visual feedback (current render vs.\ target) and iteratively revises its structure while maintaining structural validity; steps~9--10 are omitted as changes from step~8 are negligible.
\textbf{Claude} demonstrates meaningful visual self-correction: it produces a plausible roof (called: gambrel style) by step~2 but with an incorrect ridge orientation, then progressively corrects across subsequent steps, converging to the almost right form by step~8.
\textbf{GPT-5} commits to an incorrect orientation from step~1 and fails to recover, instead collapsing into a hybrid that blends both orientations without resolving either.
\textbf{Gemini} makes minimal modifications across all 10 attempts, suggesting it does not meaningfully incorporate the visual feedback signal.
This example illustrates the key behavioral divergence in this experiment: models differ not only in final score, but in \emph{whether} they engage with feedback at all.}
\label{fig:exp3_progress}
\end{figure*}

We evaluate each model under two input conditions:
\textit{Frame}, in which the model receives a bare timber framing
image and must generate a structurally valid assembly, and
\textit{Facade}, in which the framing is occluded by finished
exterior cladding and the model must infer the underlying structure
from appearance alone. Intuitively, the \textit{Facade} condition presents a more challenging task, as it requires the model to solve an inverse problem of hallucinating hidden, load-bearing topologies based solely on superficial exterior cues.
Table~\ref{tab:main_results} reports
structural pass rates, visual and topological fidelity, and joint pass rates under three agentic generation tasks mentioned in~\ref{sec:tasks}. Results reveal four cross-cutting patterns that
together characterize the current frontier of physically grounded generative reasoning.
We detail each observation \textit{O} in turn.

\begin{tcolorbox}[colback=gray!4, colframe=black!60, arc=4pt,
                  boxrule=0.8pt, left=8pt, right=8pt,
                  top=4pt, bottom=4pt]
\textit{O1: Structural validity and visual fidelity are orthogonal.}
\end{tcolorbox}
\noindent GPT-5 leads structurally in \ProtoOS~(79.2\%) yet scores lowest
visually (0.312); Claude leads visually (0.406) but not structurally;
Gemini leads structurally under \ProtoSW~(78.5\%) while remaining
competitive visually (0.313).
Models frequently satisfy validation by converging to a generic,
physically safe assembly that ignores visual conditioning entirely.
Physical validity is not a byproduct of visual imitation, and vice versa.

\begin{tcolorbox}[colback=gray!4, colframe=black!60, arc=4pt,
                  boxrule=0.8pt, left=8pt, right=8pt,
                  top=4pt, bottom=4pt]
\textit{O2: Structural reasoning is not a monolithic capability.}
\end{tcolorbox}
\noindent Although GPT-5 leads structurally in \ProtoOS~(79.2\%), it collapses under
\ProtoSW~(33.3\%), dropping 46 points; Gemini reverses this entirely
(45.4\% $\to$ 78.5\%), gaining 33 points; Claude is comparably protocol-stable (71.6\% vs.\ 71.3\%).
We interpret this as two distinct reasoning modes: \emph{holistic}
(\eg, GPT-5 is strong single-pass synthesis, weak under incremental commitment)
and \emph{\ProtoSW} (\eg, Gemini is strongest with phase decomposition and
intermediate feedback).
\ProtoOSeq~experiment confirms this: forcing models to self-decompose a structure into sequential steps without any external phase guidance is harder, as it demands both holistic planning and incremental commitment simultaneously.

\begin{tcolorbox}[colback=gray!4, colframe=black!60, arc=4pt,
                  boxrule=0.8pt, left=8pt, right=8pt,
                  top=4pt, bottom=4pt]
\textit{O3: Some Models can see structure under the skin, some cannot.}
\end{tcolorbox}
\noindent Claude and Gemini improve structurally under Facade input
(Claude $71.6\% \to 77.9\%$; Gemini $45.4\% \to 53.9\%$);
GPT-5 declines sharply ($79.2\% \to 63.7\%$).
Gemini is the only model whose visual score also improves ($0.376 \to 0.394$),
suggesting it activates strong architectural priors when reasoning
through appearance to infer underlying structure.
GPT-5, relying on direct structural mimicry, is disrupted when
the framing signal is occluded by cladding.

\begin{tcolorbox}[colback=gray!4, colframe=black!60, arc=4pt,
                  boxrule=0.8pt, left=8pt, right=8pt,
                  top=4pt, bottom=4pt]
\textit{O4: Protocol dominates model.}
\end{tcolorbox}
\noindent The performance gap attributable to protocol choice exceeds any
cross-model difference within a fixed protocol.
Gemini swings from 45.4\% to 78.5\% structural pass rate, a
33 points gain, simply by changing from \ProtoOS~to \ProtoSW.
Crucially, Gemini \textit{overtakes} GPT-5 under \ProtoSW~despite
trailing it by 34 points under \ProtoOS, a rank reversal
impossible to predict from single-protocol evaluation alone.
This suggests that agentic scaffold design is at least as consequential
as model selection: the same model can be the weakest or the strongest
depending on how the task is structured.

\paragraph{Other Observations.}
Topological proximity does not guarantee structural validity.
Composite topological scores are nearly identical between passed and failed
structures: in \ProtoOS, passed vs.\ failed scores are
0.164 vs.\ 0.149 (Claude), 0.171 vs.\ 0.175 (Gemini), 0.143 vs.\ 0.131 (GPT-5); \textit{see Appendix for details}.
For Gemini, failed structures score \emph{higher} topologically than passing ones.
Physical constraints are \emph{discontinuous}, a single missed connection
or an over-span fails the entire structure with no partial credit while topological and visual metrics are smooth and continuous.
Because of this fundamental mismatch, standard perceptual benchmarks cannot be relied upon. They merely measure how accurate a structure \emph{looks} overall, making them insufficient proxies for evaluating whether a model actually understands true physical generative reasoning.

\subsection{Iterative Refinement with Visual Feedback }
\label{sec:exp3}

Can iterative visual feedback improve similarity while maintaining structural validity?
Starting from \ProtoOS~structural passes with Visual Fidelity $S < 0.6$ (\textit{see Appendix for reason}), we run up to 10
visual-feedback revisions with structural validation enforced at every step.
Table~\ref{tab:exp3} reports results.

Structural retention at the final attempt is 68.5\% for Claude, 73.8\%
for Gemini and 78.1\% for GPT-5, confirming visual feedback does not
catastrophically destabilize structural integrity.
Visual scores improvement is moderate but consistent: Claude gains most
(+0.033, 70.3\% of Tasks improved);
Gemini and GPT-5 gain less but still with more than half of the tasks' visual score improved.

%% file: tables/results_table.tex
\begin{table*}[t!]
\centering
\caption{\small Consolidated Performance Metrics for GPT-5, Claude Opus 4.5, and Gemini 3 Flash.
  \textit{Frame}~=~bare timber framing; \textit{Facade}~=~finished exterior.
  \textbf{Bold}~=~best per column;
  \colorbox[RGB]{220,240,220}{shaded rows}~=~best model per protocol by average across all conditions.
  \textbf{Structural Pass Rate} in $[0,1]$: passes only if all 16 validation tests clear.
  \textbf{Visual Fidelity}: average visual similarity score across 5 rendered views.
  \textbf{Topological Fidelity}: structural similarity across member counts, spatial alignment, and volumetric overlap.
  \textbf{Joint Pass Rate} in $[0,1]$: fraction satisfying \emph{both} structural and visual criteria simultaneously.}
\label{tab:main_results}
\resizebox{\textwidth}{!}{%
\begin{tabular}{ll cccc cc | cc}
\toprule
\multirow{2}{*}{\textbf{Model}}
  & \multirow{2}{*}{\textbf{Protocol}}
  & \multicolumn{2}{c}{\textbf{Structural Pass Rate}}
  & \multicolumn{2}{c}{\textbf{Visual Fidelity}}
  & \multicolumn{2}{c}{\textbf{Topological Fidelity}}
  & \multicolumn{2}{c}{\textbf{Joint Pass Rate}} \\
\cmidrule(lr){3-4}\cmidrule(lr){5-6}\cmidrule(lr){7-8}\cmidrule(lr){9-10}
& & \textit{Frame} & \textit{Facade}
  & \textit{Frame} & \textit{Facade}
  & \textit{Frame} & \textit{Facade}
  & \textit{Frame} & \textit{Facade} \\
\midrule

\multirow{3}{*}{GPT-5}
  & \ProtoOS{}   & \textbf{0.792} & 0.637 & 0.312 & 0.287 & 0.143 & 0.151 & 0.035 & 0.019 \\
  & \ProtoOSeq{} & 0.302 & 0.247 & 0.293 & 0.266 & 0.141 & 0.143 & 0.003 & 0.003 \\
  & \ProtoSW{}   & 0.333 & 0.157 & 0.179 & 0.175 & 0.178 & 0.157 & 0.008 & 0.003 \\
\midrule

\multirow{3}{*}{Claude}
  & \Best{\ProtoOS{}}   & \Best{0.716} & \Best{\textbf{0.779}} & \Best{\textbf{0.406}} & \Best{0.377} & \Best{0.164} & \Best{\textbf{0.168}} & \Best{\textbf{0.071}} & \Best{\textbf{0.064}} \\
  & \ProtoOSeq{} & 0.428 & 0.494 & 0.239 & 0.245 & 0.133 & 0.141 & 0.003 & 0.013 \\
  & \ProtoSW{}   & 0.713 & 0.737 & 0.278 & 0.291 & 0.205 & 0.190 & 0.031 & 0.022 \\
\midrule

\multirow{3}{*}{Gemini}
  & \ProtoOS{}   & 0.454 & 0.539 & 0.376 & \textbf{0.394} & 0.171 & 0.170 & 0.031 & 0.036 \\
  & \Best{\ProtoOSeq{}} & \Best{0.507} & \Best{0.376} & \Best{0.345} & \Best{0.342} & \Best{0.160} & \Best{\textbf{0.152}} & \Best{0.019} & \Best{0.013} \\
  & \Best{\ProtoSW{}}   & \Best{0.785} & \Best{0.737} & \Best{0.313} & \Best{0.282} & \Best{\textbf{0.232}} & \Best{0.211} & \Best{0.043} & \Best{0.026} \\

\bottomrule
\end{tabular}}
\end{table*}

%% file: tables/exp3_table.tex
\begin{table*}[t!]
\centering
\caption{\small Iterative refinement with visual feedback.
  Each model starts from its \ProtoOS~structural passes
  and iterates up to 10 visual-feedback revisions.
  \textit{Baseline} = oneshot output before revision;
  \textit{Final} = last attempt;
  \textit{Best} = highest-scoring attempt across all iterations.
  $\Delta$ is Final$-$Baseline mean score.
  \textit{Improved Tasks} = fraction of tasks that improved their visual fidelity score.
  \textit{Structures Retained} = fraction retaining structural validity at the final attempt.
  \textbf{Bold} = best per column.}
\label{tab:exp3}
\centering
\resizebox{0.8\linewidth}{!}{
\begin{tabular}{@{} l ccc c r r @{}}
\toprule
& \multicolumn{3}{c}{\textbf{Mean Visual Score}} & & & \\
\cmidrule(lr){2-4}
\textbf{Model}~~
  & ~~\textit{Baseline} & ~~\textit{Final} & ~~\textit{Best}
  & $\Delta$ & \textit{\% Improved Tasks} & ~~\textit{Structures\ Retained} \\
\midrule
\Best{Claude} & \Best{0.408} & \Best{\textbf{0.441}} & \Best{\textbf{0.461}} & \Best{\textbf{+0.033}} & \Best{\textbf{70.3\%}} & \Best{68.5\%} \\
GPT-5 & 0.392 & 0.405 & 0.425 & +0.013 & 61.4\% & \textbf{78.1\%} \\
Gemini & 0.433 & 0.443 & 0.471 & +0.010 & 56.1\% & 73.8\% \\
\bottomrule
\end{tabular}
}
\end{table*}

%% file: sections/conclusion.tex

\section{Conclusion}
\label{sec:conclusion}

We introduced DreamHouse, a benchmark for \emph{physical generative reasoning} in timber-frame construction. 
Evaluating four frontier VLMs reveals that structural validity and visual plausibility are orthogonal signals. 
Models ranking similarly on standard leaderboards diverge sharply here, yielding structural success rates from 15.7\% to 79.2\% and exhibiting qualitatively distinct failure modes. 
Crucially, performance is highly sensitive to the generation protocol: a model's assigned paradigm (holistic vs.\ stepwise) often impacts correctness more than the choice of model itself. 
This demonstrates that \emph{how} a model is asked to reason about physical space matters as much as what it inherently knows.
These results expose a gap in current evaluations: the ability to maintain global structural consistency, such as load-path continuity and code-compliant spans within complex real-world building systems. 
We hope DreamHouse and its validation suite serve as a concrete, verifiable target to track progress in this domain.

\noindent{\textit{Limitations and future work.}}
High computational costs for executing validation scripts at scale currently limit our evaluation to closed-source frontier models. 
Future work should extend our physical constraint vocabulary to encompass concrete, steel, or multi-material assemblies. 
Furthermore, while we primarily evaluate the \emph{output} generation pipeline, exploring the \emph{input} end is vital, identifying visual representations, supervision signals, or training objectives that enable models to proactively internalize structural grammar rather than merely recovering it through iterative feedback.

%% file: appendix.tex
\clearpage
\begin{center}
  {\Large\bfseries Appendix}
\end{center}

\noindent Appendix provides additional details on the dataset construction, evaluation protocol, model prompts, and extended experimental results that complement the main paper.

\section{Structural Validation Suite}
\subsection{Load Path (Topological Connectivity)}

\subsubsection{Setup}

Let $\mathcal{S} = (V, E)$ be an undirected graph where
$V = \{v_1, \dots, v_n\}$ is the set of structural members
and $E$ is the adjacency relation.
For each member $v_i$, define its world-space
axis-aligned bounding box (AABB):
\[
  \mathrm{AABB}_i =
    [x_i^-,x_i^+]\times[y_i^-,y_i^+]\times[z_i^-,z_i^+]
\]

\begin{definition}[Contact Relation]
Two members $v_i, v_j \in V$ are \emph{adjacent},
written $v_i \sim v_j$, iff their AABBs overlap with
tolerance $\varepsilon = 0.05\,\text{m}$, i.e.\ for each
axis $k\in\{x,y,z\}$ the true gap satisfies
\[
  \delta_k = \max\!\bigl(0,\,
    \max(a_k^-,b_k^-)-\min(a_k^+,b_k^+)\bigr)\le\varepsilon.
\]
\end{definition}

\begin{definition}[Ground Set]
$G = \{v_i \in V \mid z_i^- < 0.1\,\text{m}\}$.
\end{definition}

\begin{definition}[Support Function]
$\sigma: V \to \{0,1\}$ is the fixed point of:
\[
  \sigma(v_i) = \begin{cases}
    1 & v_i \in G \\
    1 & \exists\,v_j \sim v_i,\; \sigma(v_j)=1 \\
    0 & \text{otherwise}
  \end{cases}
\]
computed iteratively until convergence.
\end{definition}

\subsubsection{Test Criterion}
\[
  \boxed{T_1 = \textsc{Pass}
    \iff \sigma(v_i)=1 \quad \forall\,v_i \in V}
\]

\subsection{Span Limits (IRC Compliance)}

\subsubsection{Setup}
For each joist or rafter $v_i$, let $L_i$ be the clear span,
$(w_i, d_i)$ the cross-section dimensions, and $\tau=0.03$
the span tolerance. IRC look-up tables give allowable spans:
\[
  \mathcal{T}_J:(w_i,d_i)\mapsto L_J^*, \quad
  \mathcal{T}_R:(w_i,d_i)\mapsto L_R^*.
\]

\subsubsection{Effective Span Reduction}
When purlins are present, rafter effective span is halved:
\[
  L_i^{\text{eff}} =
    \begin{cases} L_i/2 & \text{purlin present} \\
                  L_i   & \text{otherwise}
    \end{cases}
\]

\subsubsection{Test Criterion}
\begin{equation*}
  \boxed{\begin{aligned}
    &T_2 = \textsc{Pass} \iff\\
    &\quad\begin{cases}
      L_i \le (1{+}\tau)L_J^*       & \forall\,v_i \in \text{Joists}\\[2pt]
      L_i^{\text{eff}} \le (1{+}\tau)L_R^* & \forall\,v_i \in \text{Rafters}
    \end{cases}
  \end{aligned}}
\end{equation*}

\subsection{On-Centre Spacing (16\textquotedbl{}/24\textquotedbl{})}

\subsubsection{Setup}
Group joists sharing the same elevation and direction into
sets $\mathcal{G}_k$. Sort each group by perpendicular
position and form the spacing sequence
$s_\ell = p_{\ell+1}-p_\ell$.
Standard spacings: $\mathcal{S}=\{0.406,0.610\}\,\text{m}$.

\begin{definition}[Compliant Spacing]
$s_\ell$ is \emph{compliant} if
$\min_{s^*\in\mathcal{S}}|s_\ell - s^*| < 0.05\,\text{m}$.
\end{definition}

\subsubsection{Test Criterion}
\[
  \boxed{T_3 = \textsc{Pass}
    \iff \forall\,\mathcal{G}_k,\,\ell:\;
      s_\ell \text{ compliant} \vee s_\ell\le 0.1\,\text{m}}
\]

\subsection{Standard Lumber Dimensions}

\subsubsection{Setup}
Let $\Lambda$ be the standard nominal-dimension set
(metric actual sizes, mm):
\begin{align*}
  \Lambda = \{&(38,89),(38,140),(38,184),\\
              &(38,235),(38,286),(89,89),(140,140)\}
\end{align*}
For each framing member $v_i$, let
$\mathbf{d}_i=(\min(\Delta_i),\mathrm{med}(\Delta_i))$
be its two smallest cross-section dimensions.

\subsubsection{Test Criterion}
\begin{equation*}
  \boxed{\begin{aligned}
    T_4 &= \textsc{Pass} \iff \forall\,v_i\;
           \exists\,(w^*,d^*)\in\Lambda\!:\\
        &\quad |d_{i,1}-w^*| < 10\,\text{mm}\\
        &\quad \wedge\; |d_{i,2}-d^*| < 20\,\text{mm}
  \end{aligned}}
\end{equation*}

\subsection{Deflection (L/360 Serviceability)}

\subsubsection{Setup}
For each joist $v_i$: uniform load $w=1900\,\text{N/m}$,
$E=12\,\text{GPa}$, and $I_i = b_i h_i^3/12$, where
$b_i=\min(\Delta x_i,\Delta y_i)$ and $h_i=\Delta z_i$.
Mid-span deflection and limit:
\[
  \delta_i = \frac{5wL_i^4}{384EI_i},
  \qquad
  \delta_i^* = \frac{L_i}{360}.
\]

\subsubsection{Test Criterion}
\[
  \boxed{T_5 = \textsc{Pass}
    \iff \forall\,v_i\in\text{Joists}:\;
      \delta_i \le (1+\tau_\delta)\,\delta_i^*}
\]
where $\tau_\delta=0.08$ (8\% deflection tolerance).

\subsection{Roof Coverage}

\subsubsection{Setup}
Partition the footprint into $1\,\text{m}\times 1\,\text{m}$
cells. Let $\mathcal{F}$ be cells occupied by floor/sill
objects, and $\mathcal{R}\subseteq\mathcal{F}$ cells covered
by at least one rafter projection (margin $\mu=0.3\,\text{m}$).
Coverage ratio:
\[
  \rho = |\mathcal{R}|\,/\,|\mathcal{F}|.
\]

\subsubsection{Test Criterion}
\[
  \boxed{T_6 = \textsc{Pass} \iff \rho \ge 0.70}
\]

\subsection{Gap Detection}

\subsubsection{Setup}
Using the grid from Test~6, the gap set is
$\mathcal{Q}=\mathcal{F}\setminus\mathcal{R}$ and the
gap ratio $\gamma = |\mathcal{Q}|/|\mathcal{F}|$.

\subsubsection{Test Criterion}
\[
  \boxed{T_7 = \textsc{Pass} \iff \gamma \le 0.20}
\]

Since $\rho+\gamma=1$, $T_7$ is equivalent to $T_6$ at the
same threshold; both are retained for diagnostic granularity.

\subsection{Cantilever Limits}

\begin{figure*}[!t]
    \centering
    \includegraphics[width=1.0\linewidth]{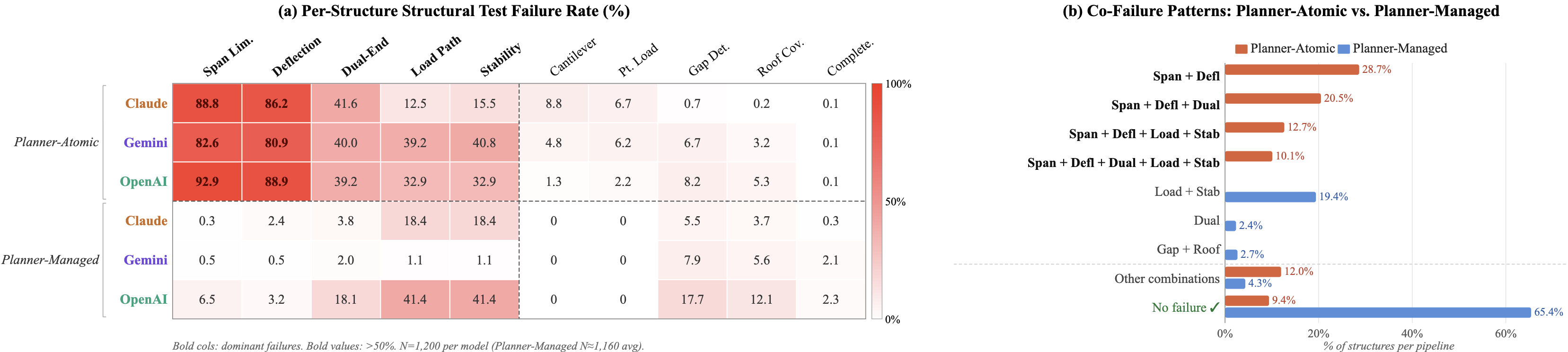}
    \caption{\small\textbf{Structural failure analysis across pipelines and models.}
\textbf{(a)} Per-structure failure rate for each of the 10 structural validation tests,
shown for the \ProtoOS~and \ProtoSW protocols across all three models
($N$\,=\,1{,}200 per model for \ProtoOS; $N$\,$\approx$\,1{,}160--1{,}197 for \ProtoSW,
excluding Phase-0 failures).
Each cell reports the percentage of structures for which that test failed
\emph{at least once} across all retry attempts.
Values are independent marginal failure rates per test and do \emph{not} sum to
100\% per row, as a single structure can fail multiple tests simultaneously.
Bold column headers indicate the five dominant failure tests.
\textbf{(b)} Co-failure pattern distribution: each structure is assigned to the
exact set of tests it failed, ranked by frequency.
Orange bars show \ProtoOS ($N$\,=\,3{,}551); blue bars show \ProtoSW
($N$\,=\,3{,}484), each normalized to their respective totals.
The top four \ProtoOS patterns share a Span\,+\,Deflection core, reflecting
near-universal geometry violations.
\ProtoSW failures are dominated by Load\,+\,Stability errors, indicating
structural connectivity issues rather than geometric ones.}
    \label{fig:supp-errormaps}
\end{figure*}

\subsubsection{Setup}
Let $\mathcal{P}$ be ground-reaching supports
($z_i^-<0.1\,\text{m}$). For each elevated sill $s$
($z_s^->1.0\,\text{m}$), let $\ell_s$ be its length and
$\mathcal{P}_s$ the nearby supports (within $c_{\max}=1.5\,\text{m}$
laterally). For long sills ($\ell_s>c_{\text{sp}}=3.0\,\text{m}$),
let $\Delta_s^{\max}$ be the maximum gap between consecutive
supports along the sill axis.

\subsubsection{Test Criterion}
\begin{equation*}
  \boxed{\begin{aligned}
    &T_8 = \textsc{Pass} \iff \forall\,s\in\text{ElevSills}:\\
    &\quad\begin{cases}
      |\mathcal{P}_s|\ge 2 \;\wedge\;
        \Delta_s^{\max}\le c_{\text{sp}} & \ell_s > c_{\text{sp}}\\[2pt]
      d(s,\mathcal{P}_s)\le c_{\max}    & \text{otherwise}
    \end{cases}
  \end{aligned}}
\end{equation*}

\subsection{Stability Score}

\subsubsection{Definition}
The \emph{Topological Stability Index} (TSI) is:
\[
  \Sigma = \frac{\sum_{v_i\in V}\sigma(v_i)}{|V|}
\]

\subsubsection{Test Criterion}
\[
  \boxed{T_9 = \textsc{Pass} \iff \Sigma \ge 1.0}
\]
$\Sigma=1.0$ means every member's load path terminates at a
grounded element. $T_9$ is strictly stronger than $T_1$:
$T_1$ checks binary grounding; $T_9$ requires the full set
$V$ to be grounded simultaneously.

\subsection{Dual-End Connection}

\subsubsection{Setup}
For each rafter or stud $v_i$ with height
$h_i=z_i^+-z_i^-\ge 0.3\,\text{m}$, define connection zones:
\[
  Z_i^{\text{bot}} = [z_i^-,\,z_i^-+\alpha h_i],\quad
  Z_i^{\text{top}} = [z_i^+-\alpha h_i,\,z_i^+]
\]
where $\alpha=0.20$.

\begin{definition}[Zone Connection]
$v_j$ \emph{connects} to zone $Z_i^{\text{bot}}$
(resp.\ $Z_i^{\text{top}}$) of $v_i$ if:
(i)~$v_j$ overlaps $v_i$ in $xy$ with tolerance
$\varepsilon_c=0.10\,\text{m}$, and
(ii)~$[z_j^-,z_j^+]$ intersects that zone with tolerance
$\varepsilon_c$.
\end{definition}

Let $\phi_i^{\text{bot}},\phi_i^{\text{top}}\in\{0,1\}$
indicate bottom/top zone connections.

\subsubsection{Test Criterion}
\begin{equation*}
  \boxed{\begin{aligned}
    &T_{10} = \textsc{Pass} \iff
      \forall\,v_i\in\text{Rafters}\cup\text{Studs}:\\
    &\quad \phi_i^{\text{bot}}=1 \;\wedge\; \phi_i^{\text{top}}=1
  \end{aligned}}
\end{equation*}

A rafter missing its top connection (ridge/collar) has a free end and may rotate under load (\emph{hinge failure}). A stud missing its bottom connection (sole plate) is a \emph{floating column}.

\begin{figure*}[htb!]
    \centering
    \includegraphics[width=0.5\linewidth]{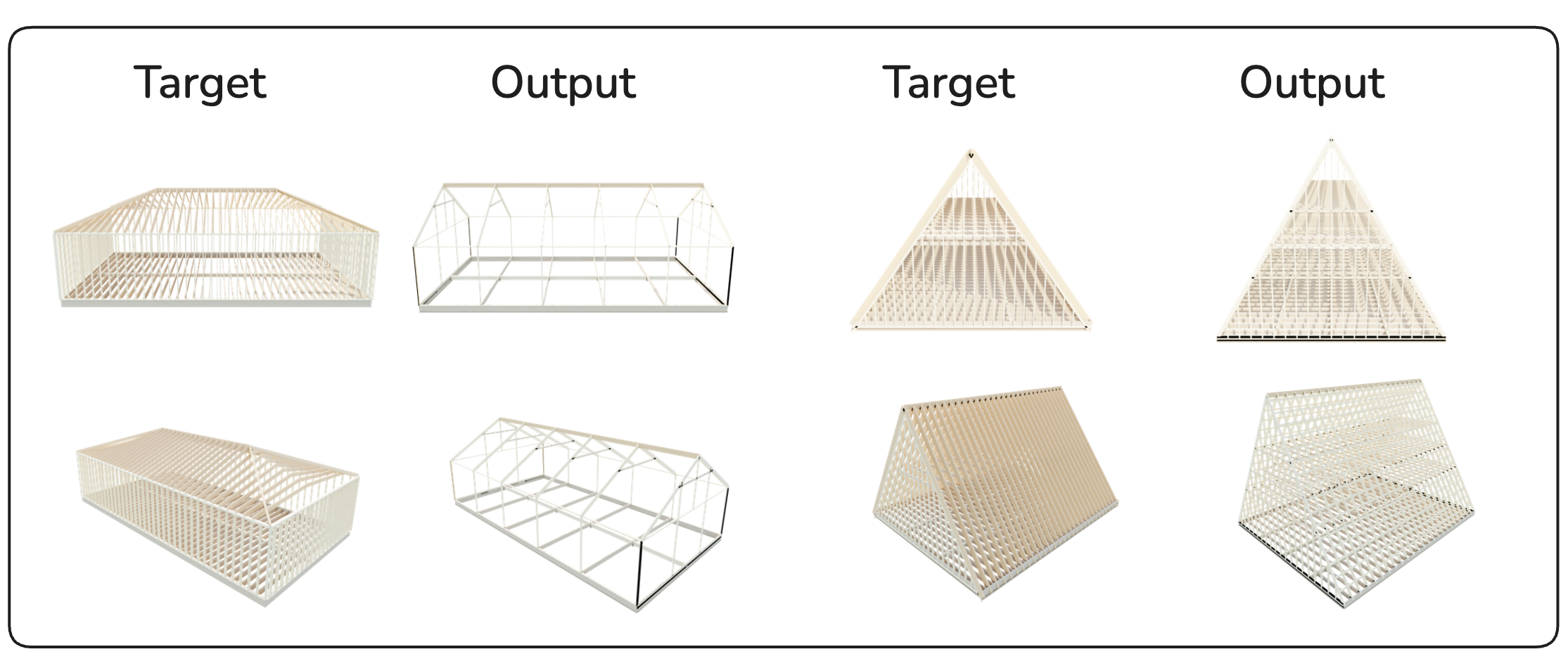}
    \caption{\small\textbf{Typical failure modes of Claude 4.5 opus (\ProtoOSeq).}
Each pair shows a ground-truth target structure (left) alongside Claude's generated output (right)
for configurations \texttt{RN\_01\_0035} (left pair) and \texttt{AF\_05\_0713} (right pair).
Claude's outputs exhibit two characteristic failure patterns:
\texttt{RN\_01\_0035} produces an overly sparse frame with missing members,
while \texttt{AF\_05\_0713} produces an overly dense packing of repeated elements.
Both reflect insufficient adherence to IRC structural rules,
the model optimizes for visual plausibility rather than code-compliant member placement,
resulting in structures that fail span limit and deflection validation
despite appearing superficially similar to the target geometry.}
    \label{fig:typical_claude_hybrid}
\end{figure*}

\begin{figure*}[htb!]
    \centering
    \includegraphics[width=0.5\linewidth]{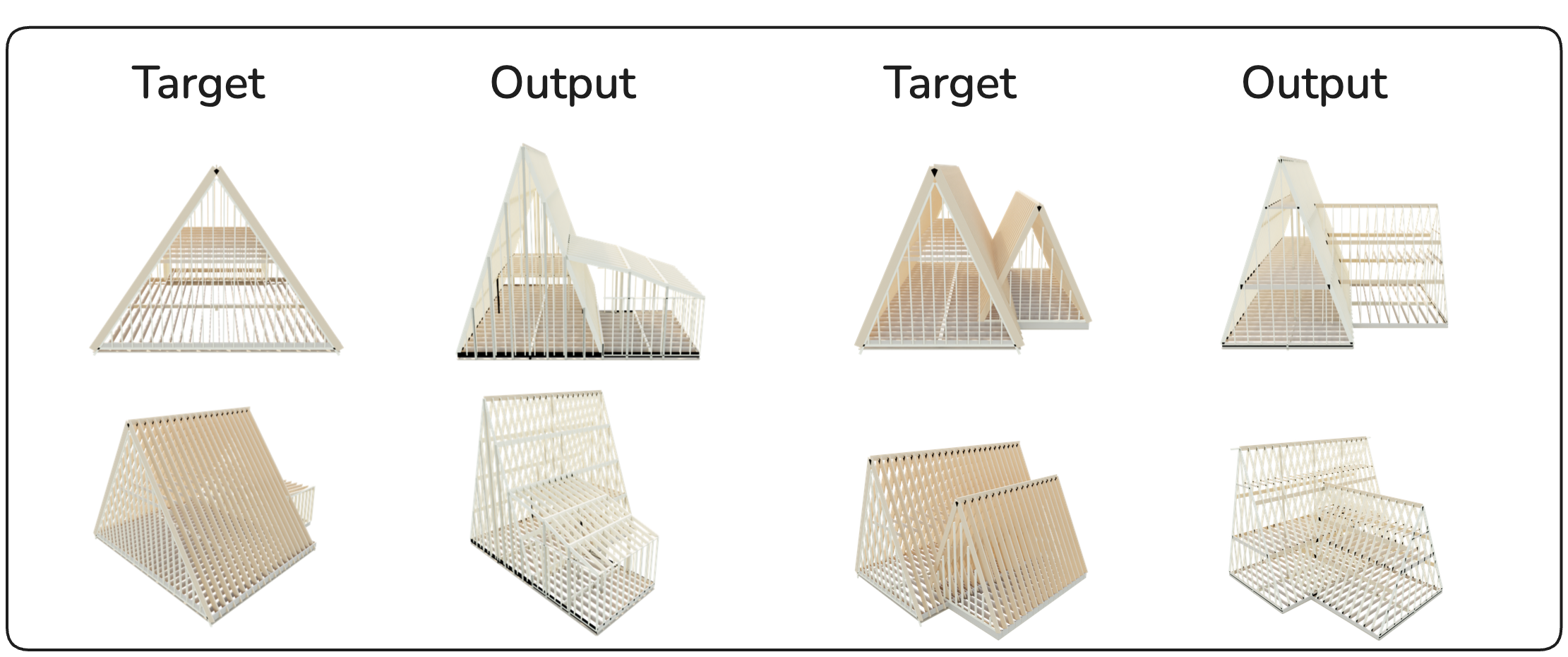}
    \caption{\small\textbf{Representative failure modes of Gemini-3-flash on A-frame structures under
the \ProtoOSeq{} protocol}. Each pair shows the target structure (left) and the
model output (right).
Left pair \texttt{AF\_03\_0392}: Gemini correctly identifies the A-frame
geometry and rafter count but misplaces the side wing, attaching it to the
wrong elevation face and inverting the attachment axis, producing a structure
that passes visual inspection from one view but fails load-path and dual-end
connection tests.
Right pair \texttt{AF\_04\_0460}: The ridge direction is rotated
$90^\circ$ relative to the target, causing all rafters to span the wrong
axis; the resulting assembly passes geometric tests ($T_3$, $T_4$) but fails
roof coverage ($T_6$) and gap detection ($T_7$) entirely.
These errors illustrate a systematic failure mode: Gemini recovers
\emph{what} structural elements are present but not \emph{where} they are
oriented, consistent with $O3$: the model attends to member topology but
not to the 3-D spatial relationships encoded in the multi-view input.}
    \label{fig:typical_gemini_hybrid}
\end{figure*}

\begin{figure*}[htb!]
    \centering
    \includegraphics[width=0.5\linewidth]{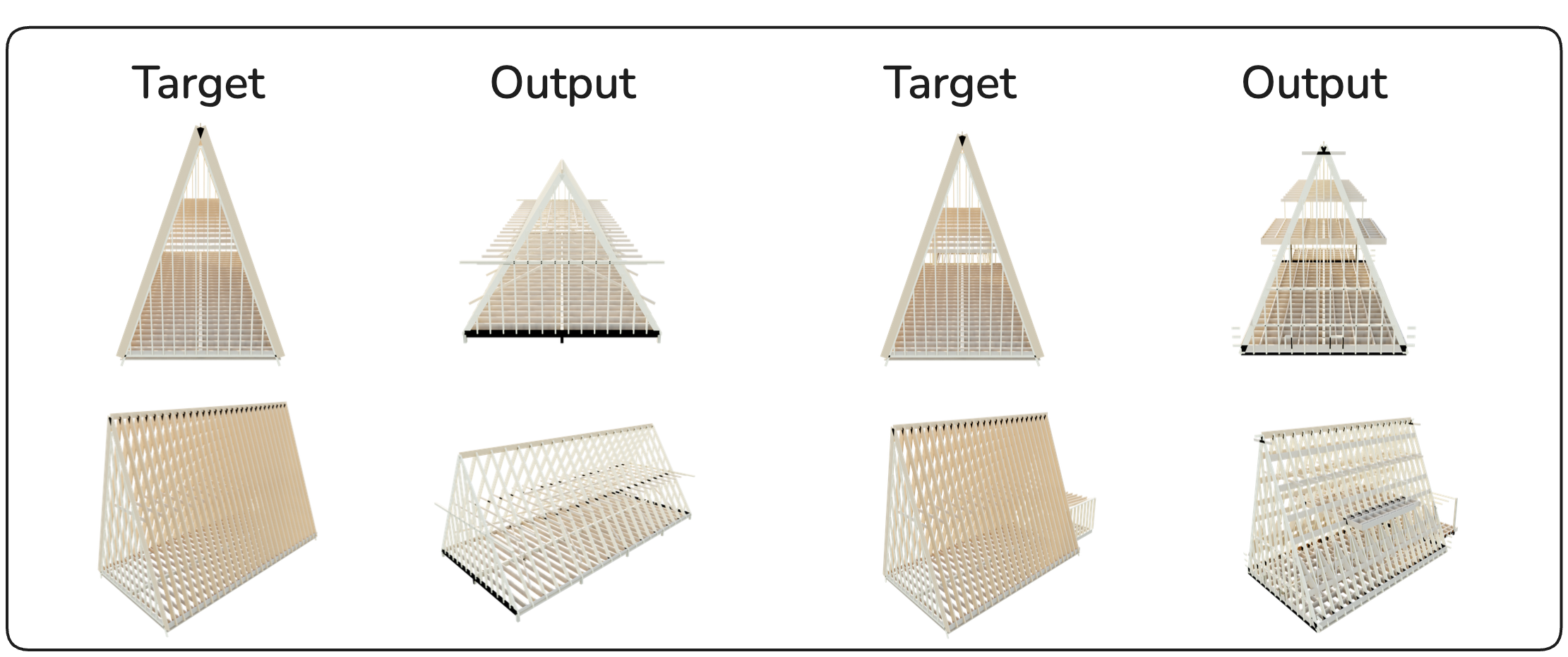}
    \caption{\small\textbf{Representative failure modes of GPT-5 on A-frame structures under
the \ProtoOSeq{} protocol.} Each pair shows the target structure (left) and the
model output (right).
Left pair \texttt{AF\_02\_0182}: GPT-5 reproduces the correct rafter count
and overall A-frame silhouette but generates collar ties and floor joists that
extend beyond the structure's bounding envelope, with members protruding
outside the intended footprint. The assembly fails dual-end connection ($T_{10}$)
and gap detection ($T_7$) as a result.
Right pair \texttt{AF\_03\_0318}: A similar pattern: the rafter and joist
layout is topologically correct but spatially unconstrained, members
overshoot the ridge and sill attachment zones, leaving floating ends
unconnected to the load path.
Both failures share a common root cause: GPT-5 reasons about \emph{which}
members to place but not about the spatial envelope that bounds them,
treating member generation as an independent local decision rather than a
globally constrained packing problem. This geometric incapability, the
inability to enforce that every member end terminates within a valid
connection zone, is the dominant failure signature of GPT-5 across
A-frame styles and is consistent with $O2$: structural reasoning is not
monolithic, and spatial constraint satisfaction is a capability that does
not co-occur with topological correctness.}
    \label{fig:typical_openai_hybrid}
\end{figure*}

\subsection*{Summary Table}

\begin{table*}[h!]
\centering
\renewcommand{\arraystretch}{1.35}
\setlength{\tabcolsep}{20pt}
\resizebox{0.8\linewidth}{!}{
\begin{tabular}{clll}
\toprule
\textbf{Test} & \textbf{Name} & \textbf{Pass Condition} & \textbf{Category} \\
\midrule
$T_1$ & Load Path & $\sigma(v_i)=1\;\forall i$ & IRC / Structural \\
$T_2$ & Span Limits & $L_i^{\text{eff}} \le (1+\tau)L^*$ & IRC / Structural \\
$T_3$ & O.C.\ Spacing & $s_\ell \in \mathcal{S} \pm 0.05\,\text{m}$ & IRC / Geometric \\
$T_4$ & Std.\ Dimensions & $(w_i,d_i) \in \Lambda$ & Geometric \\
$T_5$ & Deflection L/360 & $\delta_i \le (1+\tau_\delta)\,L_i/360$ & Physics \\
$T_6$ & Roof Coverage & $\rho \ge 0.70$ & Geometric \\
$T_7$ & Gap Detection & $\gamma \le 0.20$ & Geometric \\
$T_8$ & Cantilever Limits & Support spacing $\le 3.0\,\text{m}$ & Structural \\
$T_9$ & Stability Score & $\Sigma = 1.0$ & Topological \\
$T_{10}$ & Dual-End Connection & $\phi_i^{\text{bot}}=\phi_i^{\text{top}}=1\;\forall i$ & Structural \\
\bottomrule
\end{tabular}}
\end{table*}



\subsection{Level of Detail 350 and the DreamHouse Validation Standard}

\paragraph{What LoD 350 Requires.}
The BIM Forum's Level of Development specification defines \emph{Level of Detail 350} (LoD 350) as the threshold at which a model element is sufficiently detailed for \emph{construction coordination}~\cite{bimforum2023lod}. Concretely, LoD 350 requires that every structural member be modeled with:
\begin{enumerate}[label=(\roman*), noitemsep]
    \item \textbf{Accurate geometry} --- cross-section dimensions matching nominal fabrication tolerances ($\le 3\,\text{mm}$ deviation from specification);
    \item \textbf{Correct spatial position} --- no hard clashes, defined as two members occupying the same physical volume with overlap exceeding construction tolerance;
    \item \textbf{Interface definition} --- each member must explicitly represent the connections and clearances needed by adjacent trades (e.g.\ fastener zones, bearing seats); and
    \item \textbf{Load-path integrity} --- the assembly must support a credible load transfer hierarchy from applied loads to the foundation without floating or unsupported members.
\end{enumerate}
Below LoD 350 (i.e.\ LoD 200--300), members may be represented as ``placeholders'' with approximate geometry; above it (LoD 400), full fabrication detailing is included. LoD 350 is therefore the \emph{minimum standard for constructibility review} and the natural target for automated structural generation.

\paragraph{Mapping LoD 350 to Our 10-Test Battery.}
Each requirement above maps directly to one or more tests in our validation suite (Table~\ref{tab:lod350_mapping}). Requirement (i) corresponds to $T_4$ (standard lumber dimensions) and $T_2$ (span limits, which implicitly enforce cross-section adequacy). Requirement (ii) is addressed by $T_7$ (gap detection) and the clash detection module embedded in $T_9$. Requirement (iii) is enforced by $T_{10}$ (dual-end connection), which checks that every rafter and stud is restrained at both ends, the minimum condition for a mechanically complete interface. Requirement (iv) is captured most directly by $T_1$ (load path) and $T_9$ (Topological Stability Index), with $T_8$ (cantilever limits) addressing the specific case of elevated sections that satisfy local connectivity but violate global load-path continuity.

\paragraph{Why LoD 350 Is the Right Target for Physical Generative Reasoning.}
Prior work on AI-assisted architectural generation has largely operated at LoD 100--200, producing massing models or schematic floor plans that are \emph{visually plausible} but not \emph{physically constructible}~\cite{nauata2020housegan, shabani2023housediffusion}. We argue that this constitutes a fundamental gap: a model that cannot satisfy LoD 350 constraints cannot reason about the physical world, only about its appearance. DreamHouse is specifically designed to close this gap. By grounding evaluation in LoD 350 criteria and operationalizing those criteria as formal, executable tests. We provide a benchmark at which the distinction between \emph{perceptual generation} and \emph{physical generation} becomes measurable.

Concretely, passing all ten tests in our suite is a \emph{necessary} (though not sufficient) condition for LoD 350 compliance in timber-frame residential construction. Sufficiency would additionally require fastener schedules, bearing plate sizing, and anchor bolt layouts (LoD 400 territory); we treat these as out of scope for the current benchmark but note them as a natural extension.

\begin{table*}[t!]
\centering
\small
\renewcommand{\arraystretch}{1.3}
\caption{Mapping of LoD 350 constructibility requirements to the DreamHouse validation tests.}
\label{tab:lod350_mapping}
\begin{tabular}{p{3.6cm} p{2.3cm} p{5.9cm}}
\toprule
\textbf{LoD 350 Requirement} & \textbf{Tests} & \textbf{What Failure Means} \\
\midrule
(i) Accurate geometry
    & $T_4$, $T_2$, $T_5$
    & Non-standard cross-sections; spans exceeding IRC limits; serviceability failure under design load \\
(ii) No hard clashes
    & $T_7$, $T_3$
    & Members overlap or leave gaps wider than one stud bay; framing cannot be physically assembled \\
(iii) Interface definition
    & $T_{10}$, $T_6$
    & Rafters/studs free at one end (hinge failure under load); roof plane incompletely framed \\
(iv) Load-path integrity
    & $T_1$, $T_9$, $T_8$
    & Floating members; unsupported elevated sections; Topological Stability Index $<1.0$ \\
\bottomrule
\end{tabular}
\end{table*}

\section{Visual Similarity Metric and Score Choice}
\label{sec:supp-metric}

\noindent We describe the full derivation of the visual similarity score $S$,
justify the choice of alpha-weighted MSE over perceptual alternatives,
and detail the view configuration used across all experiments.

\subsection{Score Derivation and Threshold Selection}
\label{sec:supp-score-deriv}

\subsubsection{Alpha-Weighted MSE Formulation.}

Let $\hat{R}_v \in [0,1]^{H \times W \times 3}$ be a rendered RGB image of a
VLM-generated structure and $R_v \in [0,1]^{H \times W \times 3}$ be the
corresponding reference render of the ground-truth structure, both produced
from the same viewpoint $v$.  Let $m_v = \alpha^g_v \cup \alpha^r_v$ be the
union alpha mask of the generated and reference renders, restricting
comparison to non-transparent regions of either structure.
The \emph{alpha-weighted mean squared error} is:

\[
  \mathrm{MSE}_v
  = \frac{\displaystyle\sum_{p}\, m_v(p) \cdot \|\hat{R}_v(p) - R_v(p)\|^2}
         {\displaystyle\sum_{p} m_v(p)}
\]

where $p$ indexes spatial pixels and the denominator normalises by the
foreground area. The union mask penalizes missing geometry: a member absent
in the generated structure leaves unmatched non-transparent pixels in the
target without rewarding empty background agreement.

\subsubsection{Per-View Score.}

The raw MSE is mapped to a bounded similarity score via a linear
transformation calibrated so that $\mathrm{MSE}_v = 0$ gives $S_v = 1.0$
(perfect match) and $\mathrm{MSE}_v = 0.1$ gives $S_v = 0.0$ (total
failure):

\[
  S_v = \max~\!\bigl(0,\; 1 - \lambda \cdot \mathrm{MSE}_v\bigr)
\]

where $\lambda = 10$. The linear form is intentional: it avoids
the saturation behaviour of sigmoid-based mappings that would compress
differences in the high-quality regime ($S_v > 0.8$) where our benchmark
discriminates most finely between models. The aggregate visual similarity
score reported in all tables is:

\[
  S = \frac{1}{V}\sum_{v=1}^{V} S_v
\]

averaged over $V = 5$ views, consistent with Eq.~(1) of the main paper.

\subsubsection{Joint Pass Condition Across Five Views.}

A single-view score is insufficient because a structure may appear correct
from the front while having a collapsed rear wall or missing roof section.
We therefore require the score to exceed threshold $\tau$ on \emph{all}
five views simultaneously.  Let $S_v$ denote the per-view score for view
$v \in \{1,\dots,5\}$.  The binary \emph{visual pass} indicator is:

\[
  \mathbb{1}_{\mathrm{vis}}
  = \prod_{v=1}^{5} \mathbf{1}\!\left[S_v \ge \tau\right]
\]

\subsubsection{Threshold Selection: $\tau = 0.6$.}

The joint pass condition is strict by design: a structure must exceed
$\tau$ on \emph{every} view simultaneously.  We set $\tau = 0.6$ as the
lowest threshold at which the joint pass rate remains non-trivially above
zero across all models. Lower values admit structures with obvious
structural failures while providing no discriminative signal, and the joint
condition at any higher threshold collapses all models to near-zero pass
rates.  The $\tau = 0.6$ operating point thus represents the most demanding
threshold that keeps the benchmark informative across all evaluated models.

\subsection{Orthographic View Configuration}
\label{sec:supp-views}

\subsubsection{View Definitions.}

All renders use \emph{perspective projection} (focal length 32\,mm) with
camera distance scaled to the structure's bounding diagonal. The five
canonical views are:

\begin{table}[t]
\centering
\small
\renewcommand{\arraystretch}{1.05}
\caption{\small Camera configuration for the five views.$c_v$ is the distance multiplier.}
\label{tab:view_config}
\begin{tabular}{lccc}
\toprule
View & Azimuth & Elevation & $c_v$ \\
\midrule
Front       & $0^\circ$   & $12^\circ$ & 1.05 \\
Back        & $180^\circ$ & $12^\circ$ & 1.05 \\
Left        & $270^\circ$ & $12^\circ$ & 1.05 \\
Right       & $90^\circ$  & $12^\circ$ & 1.05 \\
Front-right & $45^\circ$  & $18^\circ$ & 1.10 \\
\bottomrule
\end{tabular}
\end{table}

\subsubsection{Rationale for View Selection.}

The five views were chosen to provide \emph{complementary coverage} of the
structure such that no single structural failure mode is invisible across
all views:

\begin{itemize}[noitemsep]
  \item \textbf{Views 1--4} (cardinal elevations) detect wall misalignment,
    wrong storey height, missing or misplaced openings, and incorrect roof
    pitch.
  \item \textbf{View 5} (front-right isometric) reveals three-dimensional
    failures: depth errors, missing rear framing, and roof asymmetries
    that are ambiguous in any single cardinal view.
\end{itemize}

\subsubsection{Distance Normalisation.}

For each view $v$, the camera is placed at distance

\[
  d_v = c_v \cdot D
\]

where $D = \|\mathbf{p}_{\max} - \mathbf{p}_{\min}\|_2$ is the world-space
bounding diagonal of the structure and $c_v$ is the per-view multiplier
(Table~\ref{tab:view_config}). The camera always points at the structure's
bounding-box centroid $\mathbf{b} = (\mathbf{p}_{\max} + \mathbf{p}_{\min})/2$.

Scaling by $D$ ensures that structures of different sizes, from
$7\,\text{m} \times 7\,\text{m}$ cottages to $14\,\text{m} \times 14\,\text{m}$
two-storey colonials, always fill the frame comparably, without requiring
a hard-coded minimum distance. Both the generated and reference renders use
identical camera parameters derived from their respective bounding diagonals,
so any residual scale difference is absorbed by the fixed
$512 \times 512$ resize applied before MSE computation.
\section{Bridge Benchmark: Motivating Example}
\label{sec:supp-bridge}

\begin{figure*}[!ht]
\centering
\includegraphics[width=0.95\linewidth]{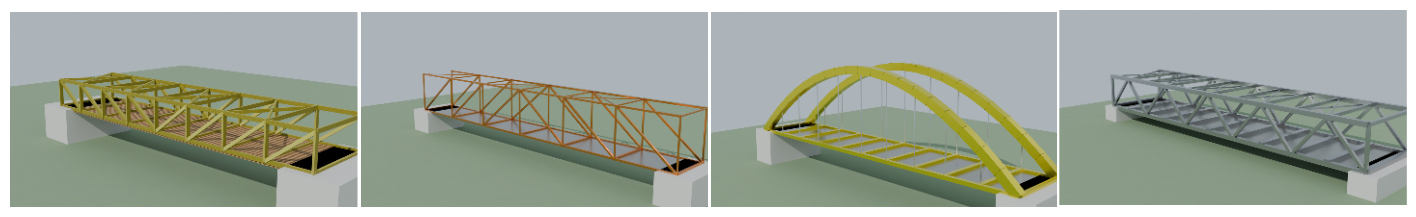}
\caption{\small Representative renders from the bridge generation pilot: glulam Howe truss (left), steel Pratt truss (center-left), steel arch (center-right), and steel Warren truss (right). Each bridge is fully parameterized by a low-dimensional \texttt{BridgeParams} dataclass whose scalar fields: \texttt{span\_length}, \texttt{truss\_height}, \texttt{num\_panels}, \texttt{deck\_width} project directly to measurable visual quantities in the render. Frontier VLMs recovered valid parameters from a single reference image in one shot with no iterative refinement, exposing the domain as a visual measurement task rather than a test of structural reasoning. This failure mode motivated the shift to timber-frame construction, where the structurally critical quantities are interior to the assembly and not directly legible from exterior renders.}
\label{fig:bridge_pilot}
\end{figure*}

Prior to settling on timber-frame construction as the testbed for
DreamHouse, we piloted a \textbf{bridge-generation} benchmark as an
alternative domain for physical generative reasoning.  The task asked
frontier VLMs to generate a structurally valid truss or arch bridge given a
reference render, outputting a \texttt{BridgeParams} dataclass that was then
validated against a 26-test suite (AASHTO LRFD~\cite{aashto2020lrfd}, AISC 360~\cite{aisc360-22}, ACI 318~\cite{aci318-19}) and
re-rendered for visual comparison.  On the surface, the domain appeared
well-suited: the validation criteria are formally codified, the geometry is
spatially unambiguous, and the renders are visually rich.

\paragraph{The Shortcut.}
We observed that frontier VLMs could recover a near-perfect
\texttt{BridgeParams} dict from a single reference image \emph{in one
shot}, requiring no iterative refinement.  The outputs passed all validation
tests and were visually indistinguishable from the ground-truth renders.
Closer inspection revealed why: the render pipeline exposes a
\emph{low-dimensional, analytically invertible} parametric space.

Concretely, the render script (\texttt{bridge\_render.py}) places the camera
at a fixed distance $d = c \cdot L$ proportional to \texttt{span\_length}
$L$, always pointed at the centroid of the bounding box.  The truss geometry
is then a deterministic function of six scalar parameters:
\texttt{span\_length}, \texttt{truss\_height}, \texttt{num\_panels},
\texttt{deck\_width}, \texttt{chord\_width}, and \texttt{arch\_rise},
each of which projects to a directly measurable visual quantity in the render:
apparent span-to-height ratio, panel count (a discrete integer legible from
the image), deck-to-span width ratio, and so on.  Furthermore, the
validation tests themselves impose \emph{tight algebraic constraints} that
further reduce degrees of freedom: the span/depth ratio must lie in $[7, 11]$
(Test 1), the panel aspect ratio in $[0.65, 1.40]$ (Test 6), the arch rise
ratio in $[0.10, 0.40]$ (Test 7).  A model that can read two distances from
an image and solve a pair of inequalities can satisfy these tests without any
structural understanding.

In short, the bridge benchmark collapsed into a \emph{visual measurement
task}: estimate a handful of lengths from the image, verify they satisfy
publicly available code ratios, and instantiate the dataclass.  No reasoning
about member connectivity, load path topology, or construction hierarchy was
required.

\paragraph{Why Timber Framing Forecloses This Shortcut.}
Timber-frame residential construction lacks a low-dimensional parametric
representation that is directly legible from exterior renders.  The
structurally critical quantities in DreamHouse: member spacing,
connection zone assignment, load-path topology ($\sigma(v_i) = 1\ \forall
i$), dual-end restraint, and IRC span compliance are \emph{interior
properties} of the framing assembly that are partially or wholly occluded in
any single render.  A wall shifted 0.5\,m or a joist missing its end
connection looks nearly identical from the outside but fails multiple
validation tests.  No analytic inversion of the render exists because the
mapping from framing topology to exterior appearance is many-to-one and
non-injective: different assemblies produce indistinguishable renders while
differing in load path, cantilever adequacy, and structural stability.

This is the key asymmetry between the two domains.  Bridge geometry is
\emph{exoskeletal} the structural members are the exterior, so visual
similarity and structural correctness are tightly coupled.  Timber framing is
\emph{endoskeletal}, the structure is concealed behind cladding
so the two are largely decoupled.  DreamHouse is specifically designed to
exploit this decoupling: a model cannot achieve high structural pass rates by
visual pattern matching alone, and the 7.1\% joint pass rate observed across
all frontier models (main paper, Table~2) confirms that the shortcut
available in the bridge domain is not available here.


\begin{figure*}[h!]
    \centering
    \includegraphics[width=0.9\linewidth]{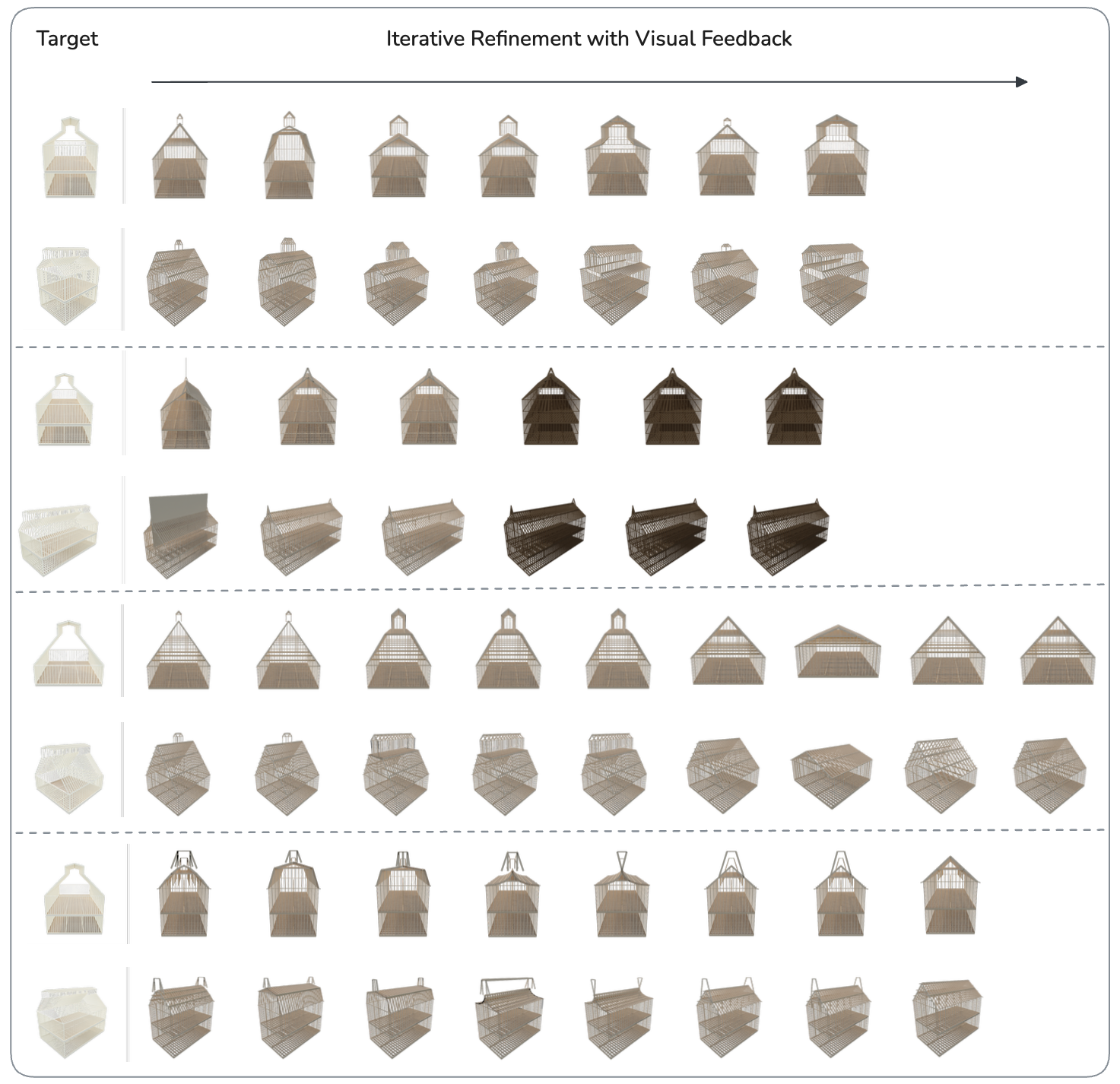}
    \caption{\small Iterative Refinement with Visual Feedback trajectories for Claude 4.5 opus on four similar
barn-style structures. Each pair of rows shows the target (leftmost, lighter color) followed
by successive iterations (left to right, darker). Dashed horizontal lines
separate groups of structurally distinct targets. Within each group, targets
share the same architectural style and similar proportions, yet Claude
consistently pursues \emph{different} refinement strategies across instances:
some trajectories make large structural revisions between iterations (roof
skeleton replaced); others converge quickly and make only
minor adjustments; others oscillate without monotonic improvement. This
variability within a homogeneous style group reveals that Claude's
behavior is not driven by a stable visual error signal, it does not
reliably identify the same class of discrepancy across similar targets and
apply a consistent corrective action. Rather, each iteration represents an
independent re-interpretation of the feedback, which can move the output
toward or away from the target depending on which aspect of the render the
model attends to. This is consistent with $O1$ and $O3$: visual similarity and
structural validity are orthogonal, and the ability to \emph{see} a
structural error in a rendered image does not imply the ability to
\emph{correct} it in the next generation.}
\label{fig:supp-qualitative-claude-vfr}
\end{figure*}

\begin{figure*}[hbt!]
    \centering
    \includegraphics[width=0.9\linewidth]{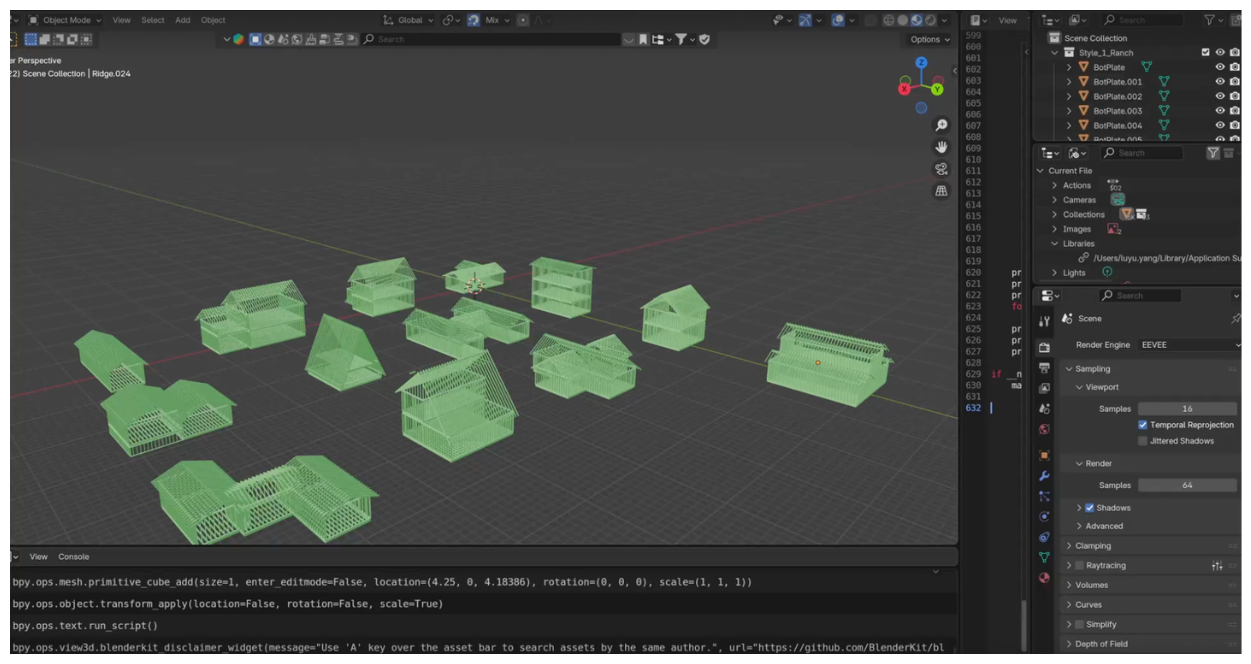}
    \caption{\small The 13 architectural style archetypes visualized in Blender,
shown as wireframe meshes in the 3D viewport. Each archetype defines
the canonical massing, roof type, section layout, and story count for
one style category in the DreamHouse dataset. The procedural
generation pipeline instantiates each archetype across a range of lot
sizes, proportions, and complexity levels to produce the 26,000+
structure dataset. Structures are rendered headless via the Blender
Python API (\texttt{bpy}); the console output visible at the bottom
shows a representative member placement call during procedural
construction.}
\label{fig:supp-blender-archetypes}
\end{figure*}

\section{Extended Experimental Results}
\label{sec:supp-results}

\subsection{Stepwise with Per-Step Visual Feedback}
\label{sec:supp-visual-feedback}

We evaluated a variant of \ProtoSW{} in which the rendered output is
compared against the target after each construction step and visual
similarity feedback is appended to the prompt before the next step.  Table~\ref{tab:visual_feedback} reports structural pass
rate, mean visual fidelity, and joint pass rate: the
fraction of structures that simultaneously pass all structural tests
\emph{and} achieve $S \ge 0.6$ for both conditions.

\noindent\textbf{Joint pass rate.}
Visual feedback more than doubles the joint pass rate for Claude
($+205\%$, $3.08\% \to 9.40\%$) and Gemini ($+105\%$,
$4.34\% \to 8.90\%$), while GPT-5 shows only a marginal improvement
($+23\%$, $0.75\% \to 0.92\%$).  The joint metric is the most
demanding criterion in our evaluation. It requires the output to be
simultaneously safe and visually faithful and the large gains for
Claude and Gemini indicate that per-step visual grounding helps models
resolve the orthogonality between structural validity and visual
similarity identified in O1.

\noindent\textbf{Structural vs.\ visual trade-off.}
The two objectives do not always improve together under visual feedback.
Claude gains on both axes: structural pass rate increases by $+6.67$
percentage points and Visual Fidelity increases by $+3.7\%$.  Gemini,
however, trades a small structural regression ($-4.45$ pp) for a large
visual gain: the fraction of structures passing $S \ge 0.6$ nearly
triples ($5.4\% \to 13.3\%$).  GPT-5 declines slightly on both axes,
suggesting that visual feedback does not compensate for its underlying
spatial constraint failures identified in
Figure~\ref{fig:typical_openai_hybrid}.

\noindent\textbf{Threshold crossing.}
The most obvious effect of visual feedback is at the $S \ge 0.6$
threshold: all three models show larger relative gains in visual pass
rate than in Visual Fidelity, indicating that feedback helps a
meaningful subset of structures cross the quality threshold even when
the overall mean shifts little.  This is consistent with visual
feedback acting as a targeted correction signal rather than a uniform
quality lift.



\begin{table*}[t!]
\centering
\small
\renewcommand{\arraystretch}{1.3}
\caption{Stepwise (\ProtoSW{}) \textbf{without} per-step visual feedback vs. \textbf{with} in \textit{Frame} (timber framing as input). Joint pass rate is the primary metric: structures
that are both structurally valid and visually similar ($S \ge 0.6$).}
\label{tab:visual_feedback}
\resizebox{0.8\linewidth}{!}{
\begin{tabular}{@{} llcccc @{}}
\toprule
\textbf{Model} & \textbf{Protocol} &
\textbf{Struct.\ Pass} & \textbf{Visual Fidelity} &
\textbf{Visual Pass ($\ge$0.6)} & \textbf{Joint Pass} \\
\midrule
\multirow{2}{*}{GPT-5}
  & \ProtoSW{}            & 0.333 & 0.179 & 0.022 & 0.008  \\
  & + visual feedback     & 0.317 & 0.173 & 0.030 & \Best{0.009}  \\
\midrule
\multirow{2}{*}{Claude}
  & \ProtoSW{}            & 0.713 & 0.278 & 0.043 & 0.031  \\
  & + visual feedback     & 0.780 & 0.288 & 0.122 & \Best{0.094}  \\
\midrule
\multirow{2}{*}{Gemini}
  & \ProtoSW{}            & 0.785 & 0.313 & 0.054 & 0.043  \\
  & + visual feedback     & 0.740 & 0.311 & 0.133 & \Best{0.089}  \\
\bottomrule
\end{tabular}
}
\end{table*}

\subsection{Open-Source Model Evaluation}
\label{sec:supp-opensource}

\begin{figure*}[t!]
    \centering
    \includegraphics[width=1.0\linewidth]{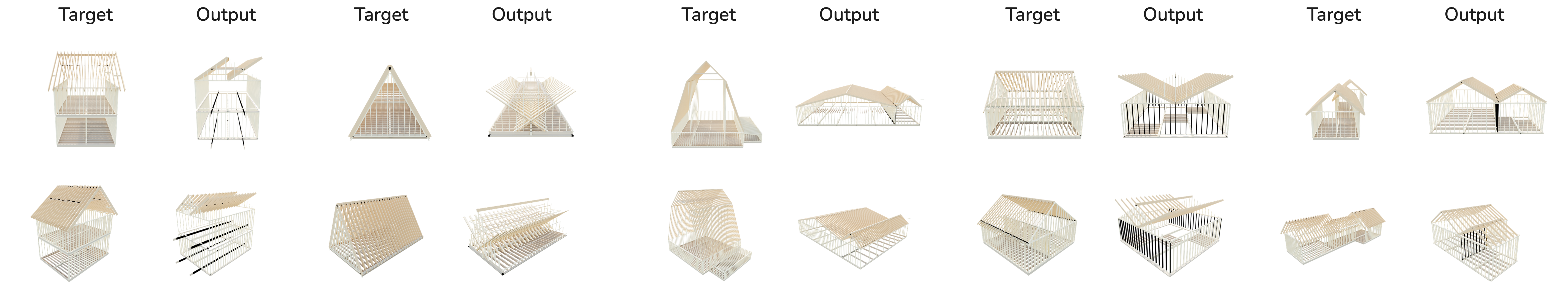}
    \caption{\small Representative outputs of Qwen3.5-397B-A17B under \ProtoSW{},
showing five target--output pairs across two views (front elevation, top;
diagonal, bottom).
Failures are more fundamental than those observed in frontier models:
\textbf{columns 1, 2, 4} show incorrect rafter orientation --- rafters
span the wrong axis entirely, producing a roof that is geometrically
inverted relative to the target;
\textbf{column 3} exhibits severe global disproportion, with the
generated structure roughly half the height and twice the footprint of
the target, indicating a failure to infer absolute scale from the
multi-view input;
\textbf{column 5} shows a collapsed depth dimension, the 3D volume
is flattened into a near-planar assembly that passes a front-view
silhouette check but has no structural depth.
Unlike the frontier model failures in
Figures~\ref{fig:typical_claude_hybrid} and~\ref{fig:typical_gemini_hybrid}, which involve spatially
misplaced but topologically coherent members, Qwen3.5 errors reflect
a lack of basic 3D spatial grounding, the model does not reliably
recover the orientation, scale, or depth of the target structure from
the provided views.}
\label{fig:supp-qualitative-qwen35}
\end{figure*}

We evaluated three open-source models under the \ProtoSW{} protocol to
assess whether the capability gap observed between frontier models
extends to publicly available weights.

\paragraph{Qwen3-VL-8B-Instruct and Qwen3-VL-30B-A3B-Instruct.}
Both models achieved a structural pass rate of 0\% across all 10
validation tests under \ProtoSW{}.  Neither model produced
syntactically valid Blender Python code reliably enough to complete
construction, and when code did execute, the resulting assemblies
failed every structural test.  We did not collect visual similarity
scores for these models as the structures were not renderable.

\paragraph{Kimi K2.5.}
We additionally ran a small pilot evaluation of Kimi K2.5 under
\ProtoSW{} on a random subset of structures and observed a structural
pass rate of 0\%, consistent with the Qwen3-VL results.  Given that
three independent open-source model families: spanning dense and
sparse MoE architectures, and parameter counts from 8B to 397B
all failed to produce analyzable structural outputs, we discontinued
the open-source evaluation track.  The benchmark in its current form
requires code generation and spatial reasoning capabilities that appear
to be frontier-only, and we leave open-source model improvement as
future work.

\paragraph{Qwen3.5-397B-A17B.}
We subsequently evaluated Qwen3.5-397B-A17B~\cite{qwen35}, a sparse
mixture-of-experts model with 17B active parameters, accessed via the
Novita AI OpenAI-compatible endpoint under the same \ProtoSW{}
hyperparameters used for frontier models (5 retries per step, 30
global retries, 5-step history truncation, $\tau_{\mathrm{retry}} =
0.4$). Visual similarity checks were disabled for this run
(\texttt{require\_visual\_similarity: false}) as the model's code
reliability was uncertain.

Out of 1{,}200 structures, 247 (20.6\%) produced renderable outputs.
The structural pass rate on this subset was 20.6\%.  Mean visual
similarity was $\bar{S} = 0.261$ (median 0.220), well below all
frontier models. Per-view scores were lowest for the back view
($S_2 = 0.167$) and highest for the front-right diagonal ($S_5 =
0.418$), consistent with the model producing a plausible front
silhouette while failing to close the rear framing.  Joint visual pass
rates collapsed rapidly with threshold: 15.4\% at $S \ge 0.3$, 10.1\%
at $S \ge 0.4$, and 1.6\% at $S \ge 0.6$.

These results confirm that the DreamHouse tasks remain
out-of-reach for current open-source models even at the 397B scale,
and that the benchmark discriminates meaningfully between frontier and
non-frontier capability levels. The 79.4\% non-renderable rate for
Qwen3.5 versus near-zero for GPT-5, Claude, and Gemini, further
highlights that reliable Blender Python code generation is itself a
frontier capability.

\subsection{Structural Error Map Visualizations}
\label{sec:supp-errormaps}

Figure~\ref{fig:supp-errormaps} provides a fine-grained breakdown of
\emph{where} models fail structurally, complementing the aggregate pass rates
reported in the main paper.  Panel~(a) reports per-test marginal failure rates
across the three protocols\ProtoOS{}, \ProtoOSeq{}) and \ProtoSW{}; panel~(b) shows the distribution of co-failure patterns,
the exact sets of tests that fail together on the same structure.

\paragraph{Dominant failure modes differ by protocol (O4).}
Under \ProtoOS protocols, failures are concentrated in the geometry cluster:
Span Limits ($T_2$), Deflection ($T_5$), and O.C.\ Spacing ($T_3$) account
for the majority of per-test failures, and the top co-failure patterns in
panel~(b) all share a Span\,+\,Deflection core.  This reflects a systematic
tendency to generate members with plausible proportions but incorrect absolute
dimensions, the structure \emph{looks} right but fails quantitative IRC
limits.  Under Stepwise, the failure profile shifts sharply toward the
connectivity cluster: Load Path ($T_1$) and Stability Score ($T_9$) dominate,
with co-failure patterns driven by Load\,+\,Stability errors.  This is
consistent with O4 (protocol dominates model): the stepwise interface
decomposes the task into member-by-member decisions that accumulate
connectivity errors even when individual members are geometrically well-formed.

\paragraph{Structural reasoning is not monolithic (O2).}
The per-test failure rates in panel~(a) reveal that no model fails uniformly
across all tests.  Models that perform well on geometric tests ($T_3$, $T_4$)
frequently fail load-path tests ($T_1$, $T_9$), and vice versa.  This
decomposition supports O2: structural competence is multi-dimensional, and a
model's overall pass rate conflates distinct capabilities: dimensional
reasoning, topological connectivity, and IRC code compliance that need not
co-occur.  The five dominant failure tests (bold column headers in panel~(a))
account for the large majority of all failures, suggesting that targeted
improvement on a small subset of structural skills would disproportionately
raise the joint pass rate.

\paragraph{Visual fidelity does not predict structural validity (O1).}
The co-failure patterns in panel~(b) reveal that many structures with high
visual similarity scores $S$ nevertheless fail multiple structural tests
simultaneously.  The Span\,+\,Deflection co-failure pattern, the most
common under \ProtoOS~produces renders that are visually near-identical to
the ground truth (correct proportions, correct topology) while failing
quantitative structural criteria.  This is the empirical basis for O1:
visual similarity and structural validity are orthogonal axes, and a benchmark
that evaluates only $S$ would misclassify the majority of these failures as
successes.

\paragraph{Some models see structure under the skin (O3).}
Across both panels, Gemini shows a distinctively lower rate of Load Path
($T_1$) and Stability ($T_9$) failures relative to GPT-5 and Claude under
the Stepwise protocol, even though its geometric failure rates are comparable.
This asymmetry, connectivity competence without geometric precision,
underlies O3: it is better at maintaining
grounded load paths member-by-member, even when individual member dimensions
are imprecise.



\subsection{Additional Qualitative Results}
\label{sec:supp-qualitative}

Figures~\ref{fig:typical_claude_hybrid},~\ref{fig:typical_gemini_hybrid} and~\ref{fig:typical_openai_hybrid}, 
show representative one-shot failures for Claude 4.5 opus, Gemini-3-flash and GPT-5 respectively.\\
Figure~\ref{fig:supp-qualitative-claude-vfr} shows iterative refinement with visual feedback trajectories for
Claude 4.5 opus across four similar barn-style structures, illustrating the high
iteration-to-iteration variance in refinement strategy even within a
homogeneous style group.



\section{Full Prompt Examples}
\label{sec:supp-prompts}

\subsection{Prompt 0: Initial Generation Prompt}
\label{sec:supp-prompt0}

Prompt~0 is the system-level instruction injected at the start of every
\ProtoOS{} and \ProtoSW{} pipeline call.  It specifies the task, the
Blender coordinate system, the structural member taxonomy, construction
dependency rules, and the output JSON schema.  The prompt is identical
across all three models; only the five rendered view images and four
scalar context fields (\texttt{lot\_size}, \texttt{stories},
\texttt{roof\_type}, \texttt{complexity}) are injected per sample.
We present the prompt in four parts.

\paragraph{Part 1: Task Definition and Building Context.}

The opening section states the task and injects per-sample metadata.
The model is asked to produce a structured JSON construction plan from
five rendered views.  Lot size, story count, roof type, and complexity
are injected verbatim from the ground-truth record and must appear
exactly in the model's output (the validator rejects plans whose
\texttt{lot\_size} fields do not match to two decimal places).

\begin{tcolorbox}[
  fontupper=\footnotesize\ttfamily,
  breakable,
  colback=gray!6,
  colframe=gray!40,
  title={\footnotesize Prompt~0 --- Part~1: \\Task Definition and Building Context},
]
\# Timber Frame Construction Planning\\
\\
\#\# Your Task\\
Create a detailed construction plan for the timber frame structure\\
shown in the images. Your plan must specify:\\
~~1. Section decomposition\\
~~2. Construction phases (foundation -> floor -> walls -> roof)\\
~~3. Step-by-step build order with dependencies\\
~~4. Classification of each step as "safe" or "critical"\\
~~5. Member types and approximate counts for each step\\
\\
\#\# Building Context  \textit{[injected per sample]}\\
- Lot Size: \{width\}m x \{depth\}m (area: \{area\}m\^{}2)\\
~~CRITICAL: Use EXACT values. Do NOT recalculate area.\\
- Stories: \{N\}\\
- Roof Type: \{gable|hip|gambrel|shed\}\\
- Complexity: \{simple|moderate|complex\}\\
\\
\#\# Images Provided (5 views, in this exact order)\\
Image 1: FRONT VIEW\\
Image 2: BACK VIEW\\
Image 3: LEFT VIEW\\
Image 4: RIGHT VIEW\\
Image 5: FRONT\_RIGHT VIEW (diagonal)
\end{tcolorbox}

\paragraph{Part 2: Coordinate System and View Definitions.}

The second section specifies the Blender world coordinate system and
provides a per-view camera table so that the model can correctly
interpret left/right/front/back directions when reading pixel-space
measurements from each image.  This is necessary because the
\texttt{back} and \texttt{right} views are mirrored relative to the
\texttt{front} view along the $x$-axis.

\begin{tcolorbox}[
  fontupper=\footnotesize\ttfamily,
  breakable,
  colback=gray!6,
  colframe=gray!40,
  title={\footnotesize Prompt~0 --- Part~2:\\ Coordinate System and View Definitions},
]
\#\# Blender Coordinate System\\
\\
~~+Z = up,~~+X = right,~~+Y = forward/front\\
\\
Axis mapping per view:\\
- front~~: camera from -Y.~~X = left-right,~~Z = up-down\\
- back~~~: camera from +Y.~~X = right-left (mirrored),~~Z = up-down\\
- left~~~: camera from +X.~~Y = front-back,~~Z = up-down\\
- right~~: camera from -X.~~Y = back-front (mirrored),~~Z = up-down\\
\\
Section bounds use:\\
- x\_min/x\_max : left-right extent\\
- y\_min/y\_max : front-back extent (negative Y = back)\\
- z\_base~~~~~~ : ground level (typically 0.0)\\
\\
\#\# How to Analyse Images\\
Step 1 (Front): count roof peaks, note step-backs, measure widths.\\
Step 2 (Sides): confirm depth, check height differences.\\
Step 3 (Diagonal): verify 3-D section relationships.\\
Step 4: cross-reference all views; use lot size as sanity check.\\
\\
Common section patterns:\\
- Main body only : rectangular footprint, one roof\\
- Main + side wing: L-shaped\\
- Main + rear wing: T-shaped\\
- Split-level~~~~: sections at different z\_base values
\end{tcolorbox}

\paragraph{Part 3: Member Taxonomy and Naming Convention.}

The third section defines the complete set of valid member types,
grouped by construction phase, together with the naming convention
enforced by the validator.  Any member whose name does not begin with
one of the listed prefixes (case-sensitive) causes a validation failure
and triggers a feedback loop.  This section also states the canonical
phase dependency chain (foundation $\to$ floor $\to$ walls $\to$ roof)
and provides typical spacing and dimension guidelines.

\begin{tcolorbox}[
  fontupper=\footnotesize\ttfamily,
  breakable,
  colback=gray!6,
  colframe=gray!40,
  title={\footnotesize Prompt~0 --- Part~3: \\Member Taxonomy and Naming Convention},
]
\#\# Structural Systems (build in this order)\\
\\
Foundation : Sill, BeamPost, Post\\
Floor~~~~~~: Rim, Joist, CenterBeam\\
Walls~~~~~~: SolePlate, TopPlate, Stud, GableStud,\\
~~~~~~~~~~~~~Header, King, Trimmer, Cripple\\
Roof~~~~~~~: Ridge, Rafter, Collar, Lookout, Purlin\\
\\
CRITICAL naming rule: member names must START WITH the prefix\\
(case-sensitive). Valid: "Sill\_front", "Joist\_floor1\_001"\\
Invalid: "beam\_001", "vertical\_member", "roof\_piece"\\
\\
Phase dependencies:\\
- Foundation complete before floor\\
- Floor complete before walls\\
- Walls complete before roof\\
- Multi-story: each story needs its own floor system\\
~~Story 1 floor at z~=~0.3m, Story 2 at z~=~3.0m, Story 3 at z~=~5.7m\\
\\
Typical dimensions (guidelines, not strict):\\
- Stud spacing~~~: 0.4m (16") on centre\\
- Joist spacing~~: 0.4m on centre\\
- Rafter spacing : 0.6m on centre\\
- Story height~~~: 2.7m floor-to-floor\\
\\
Step classification:\\
CRITICAL: phase transitions, new sections, elevated construction\\
SAFE~~~~: adding parallel members (more joists, more studs)
\end{tcolorbox}

\paragraph{Part 4: Output JSON Schema.}

The final section specifies the exact JSON schema the model must
produce.  The validator parses this output and checks: (i) that
\texttt{lot\_size} fields match the injected values exactly; (ii) that
all member-type strings match the taxonomy in Part~3; (iii) that
\texttt{depends\_on} indices form a valid directed acyclic graph (DAG) with no cycles; and (iv)
that section bounds are consistent with the declared \texttt{lot\_size}.
A condensed excerpt of the schema is shown below; the full output
includes \texttt{expected\_member\_counts} for all member types.

\begin{tcolorbox}[
  fontupper=\footnotesize\ttfamily,
  breakable,
  colback=gray!6,
  colframe=gray!40,
  title={\footnotesize Prompt~0 --- Part~4: \\Output JSON Schema (excerpt)},
]
\{\\
~~"analysis": \{\\
~~~~"description": str,\\
~~~~"stories": int,\\
~~~~"sections": [str, ...],\\
~~~~"roof\_type": "gable|hip|gambrel|shed",\\
~~~~"complexity": "simple|moderate|complex",\\
~~~~"lot\_size": \{ "width": float, "depth": float, "area": float \}\\
~~\},\\
~~"sections": [\{\\
~~~~"name": str,\\
~~~~"bounds": \{\\
~~~~~~"x\_min": float, "x\_max": float,\\
~~~~~~"y\_min": float, "y\_max": float, "z\_base": float\\
~~~~\},\\
~~~~"stories": int,\\
~~~~"systems": ["foundation","floor","walls","roof"],\\
~~~~"dependencies": [str]\\
~~\}],\\
~~"construction\_order": [\{\\
~~~~"step": int,\\
~~~~"section": str,\\
~~~~"phase":"foundation|floor|walls|roof",\\
~~~~"step\_type": "critical|safe",\\
~~~~"members": [\{ "type": str, "count": int \}],\\
~~~~"depends\_on": [int]\\
~~\}],\\
~~"expected\_member\_counts": \{ "Sill": int, "Joist": int, ... \}\\
\}
\end{tcolorbox}

\subsection{API Hyperparameters}
\label{sec:supp-api}

Table~\ref{tab:api_hyperparams} reports the API hyperparameters used
for each model--protocol combination.  All models use temperature
$= 0.0$ to ensure deterministic outputs, except GPT-5 which does not
expose a temperature parameter.  The three protocols correspond to
\ProtoOS{} (Planner-Atomic, one-shot), \ProtoOSeq{} (Planner-Reactive,
hybrid cumulative), and \ProtoSW{} (Planner-Managed, stepwise).
The iterative refinement with visual feedback pipeline uses the largest
token budget (65{,}536) to accommodate full blend-file code revisions
within a single API call.

\begin{table*}[t!]
\centering
\small
\renewcommand{\arraystretch}{1.3}
\caption{API hyperparameters per model and protocol.
$\tau_{\mathrm{retry}}$: visual similarity threshold below which the
pipeline retries the current step or full generation.
\textbf{N/A}: parameter not supported by the API.
\textbf{---}: not applicable for this protocol.}
\label{tab:api_hyperparams}
\resizebox{0.92\linewidth}{!}{
\begin{tabular}{@{} llcccccc @{}}
\toprule
\textbf{Protocol} & \textbf{Model} & \textbf{Temp.} &
\textbf{Max Tokens} & \textbf{Max Retries} & $\tau_{\mathrm{retry}}$ &
\textbf{Context History} & \textbf{Render Engine} \\
\midrule
\ProtoOS{}   & GPT-5  & N/A & 16{,}384 & 5  & 0.8 & full         & -  \\
\ProtoOS{}   & Gemini & 0.0 & 16{,}384 & 5  & 0.8 & full         & -  \\
\ProtoOS{}   & Claude & 0.0 & 16{,}384 & 5  & 0.8 & full         & -  \\
\midrule
\ProtoOSeq{} & GPT-5  & N/A & 64{,}000 & 10 & 0.6 & full         & -  \\
\ProtoOSeq{} & Gemini & 0.0 & 64{,}000 & 10 & 0.6 & full         & -  \\
\ProtoOSeq{} & Claude & 0.0 & 64{,}000 & 10 & 0.6 & full         & -  \\
\midrule
\ProtoSW{}   & GPT-5  & N/A & 16{,}384 & 5 / 30 & 0.4 & full    & -  \\
\ProtoSW{}   & Gemini & 0.0 & 16{,}384 & 5 / 30 & 0.4 & full    & -  \\
\ProtoSW{}   & Claude & 0.0 & 16{,}384 & 5 / 30 & 0.4 & last 5  & -  \\
\midrule
Iter.\ Refinement & GPT-5  & N/A & 65{,}536 & 10 & 0.6 & full   & Cycles \\
Iter.\ Refinement & Gemini & 0.0 & 65{,}536 & 10 & 0.6 & full   & Cycles \\
Iter.\ Refinement & Claude & 0.0 & 65{,}536 & 10 & 0.6 & full   & Cycles \\
\bottomrule
\end{tabular}
}
\end{table*}

\noindent\textbf{Max retries under \ProtoSW{}.}
The stepwise protocol applies two retry limits: up to 5 retries
\emph{per step} if the partial structure falls below $\tau_{\mathrm{retry}}$,
and a global cap of 30 retries across all steps per structure.

\noindent\textbf{Context history.}
All three protocols maintain full conversation history during retries:
each subsequent attempt receives all previous code outputs and
structural validation error messages.
For \ProtoOS{} and \ProtoOSeq{}, this is managed implicitly by the chat
client, which accumulates the full \texttt{messages} list across
attempts.\\
For \ProtoSW{}, history is managed by the \texttt{IterationController},
which exposes a \texttt{conversation\_history\_steps} parameter:
Gemini retains the full construction transcript (999 steps), while
GPT-5, Claude, and all other models truncate to the last 5 steps to
prevent context overflow from accumulating per-step code outputs.

\noindent\textbf{Render engine.}
The iterative refinement pipeline requires on-the-fly rendering at each
feedback iteration; Cycles is used here to produce higher-fidelity
images for the visual feedback signal.  The visual similarity scores
$S_v$ reported across all protocols are also computed from Cycles
renders to ensure a consistent comparison baseline.


\section{Benchmark Construction Details}
\label{sec:supp-dataset}

\subsection{Blender Visualization}
\label{sec:supp-blender}
Figure~\ref{fig:supp-blender-archetypes} shows the 13 architectural style
archetypes visualized in the Blender 3D viewport, which serve as the
canonical massing templates for procedural dataset generation.
The scene outliner for a representative ground-truth structure
(\texttt{RN\_01\_0119}) illustrates the taxonomy-compliant member naming
convention enforced across all 26,000+ structures in the dataset.



\subsection{Compute Environment}
\label{sec:supp-compute}

All dataset generation, structural validation, and rendering were performed
using \textbf{Blender 4.5.4 LTS} (build hash \texttt{b3efe983cc58},
commit 2025-10-27) with the \textbf{Bonsai BIM} add-on for IFC-level
structural metadata handling. Both tools are freely available open-source
software.

Structural validation (the 10-test battery described in
Section~\ref{sec:supp-metric}) is purely geometric and algebraic, it operates on member bounding boxes and graph connectivity, and
requires no GPU.  VLM API calls are likewise CPU-bound on the client
side.  GPU resources are used exclusively for Cycles rendering during
the iterative refinement pipeline, where on-the-fly photorealistic
renders are required as visual feedback at each iteration.

The pipeline is designed to be infrastructure-agnostic: the only
hard dependency is a Blender installation accessible from the command
line. No custom CUDA kernels, distributed training, or specialised
hardware are required to reproduce the benchmark.